\documentclass[12pt]{l4dc2023}
\usepackage{times}  
\usepackage{helvet}  
\usepackage{courier}  
\title{Provably Efficient Model-free RL in Leader-Follower MDP with Linear Function Approximation}
\author{%
 \Name{Arnob Ghosh} \Email{ghosh.244@osu.edu}\\
 \addr Electrical and Computer Engg.,
 The Ohio State University\\
}

\usepackage[english]{babel}
\usepackage{tikz}
\usepackage{commath}
\usepackage{breqn}
\usepackage{bbm}
\usepackage{bm}
\usepackage{algorithm}
\usepackage[noend]{algorithmic}
\usepackage{booktabs}
\usepackage{hyperref}
\newtheorem{assum}{Assumption}
\newtheorem{lem}{Lemma}
\newtheorem{goal}{Goal}
\newtheorem{cor}{Corollary}

\usepackage{enumitem}

\newcommand{\R}{\mathbb R}

\newcommand{\rD}{\mathrm{D}}

\newcommand{\rI}{\mathrm{I}}

\newcommand{\rS}{\mathrm{S}}
\newcommand{\rT}{\mathrm{T}}

\newcommand{\cA}{\mathcal A}
\newcommand{\cB}{\mathcal B}
\newcommand{\cC}{\mathcal C}

\newcommand{\cF}{\mathcal F}

\newcommand{\cL}{\mathcal L}

\newcommand{\cO}{\mathcal O}

\newcommand{\cQ}{\mathcal Q}

\newcommand{\cS}{\mathcal S}
\newcommand{\cT}{\mathcal T}

\newcommand{\cV}{\mathcal V}

\newcommand{\va}{\mathbf{a}}
\newcommand{\bI}{\mathbf{I}}
\newcommand{\bP}{\mathbbm{P}}
\newcommand{\bD}{\mathbbm{D}}
\newcommand{\bE}{\mathbbm{E}}
\newcommand{\vA}{\mathbf{A}}

\definecolor{colmb}{rgb}{0,0,.5}
\definecolor{colms}{rgb}{0,.5,0}
\definecolor{colag}{rgb}{.5,0,.5}
\definecolor{coltp}{rgb}{.5,0,0}

\newcommand{\vX}{\mathbf{X}}

\begin{document}

\maketitle
\begin{abstract}
    We consider a multi-agent episodic MDP setup where an agent (leader) takes action at each step of the episode followed by another agent (follower). The state evolution and rewards depend on the joint action pair of the leader and the follower. Such type of interactions can find applications in many domains such as smart grids, mechanism design, security, and policymaking.  We are interested in how to learn policies for both the players with provable performance guarantee under a bandit feedback setting.   We focus on a setup where both the leader and followers are {\em non-myopic}, i.e., they both seek to maximize their rewards over the entire episode and consider a linear MDP which can model continuous state-space which is very common in many RL applications.  We propose a {\em model-free} RL algorithm and show that  $\tilde{\mathcal{O}}(\sqrt{d^3H^3T})$ regret bounds can be achieved for both the leader and the follower, where $d$ is the dimension of the feature mapping, $H$ is the length of the episode, and $T$ is the total number of steps under the bandit feedback information setup. Thus, our result holds even when the number of states becomes infinite. The algorithm relies on {\em novel} adaptation of the LSVI-UCB algorithm. Specifically, we replace the standard greedy policy (as the best response) with the soft-max policy for both the leader and the follower. This turns out to be key in establishing uniform concentration bound for the value functions. To the best of our knowledge, this is the first sub-linear regret bound guarantee for the Markov games with non-myopic followers with function approximation.
\end{abstract}
\section{Introduction}
 Multi-agent Reinforcement Learning (MARL) has become important tool for decision making in a Markov game involving multiple agents. In many real-world problems, agents often have asymmetric roles. For example, one agent (leader) can act first and observing the action of the leader, the other agent (follower) reacts at each step of the MDP. This type of interaction requires two levels of thinking: the leader must reason what the follower would do in order to find its optimal decision. For example, an electric utility  company (the {\em leader}) seeks to maximize the social welfare by selecting prices at different times over a day while the users (the {\em followers}) seek to optimize their own consumption based on the prices set by the utility company. Here, the sequential decision making process can be modeled as MDP where at each period the utility company sets its price first, then, the users decide their consumption. The reward and the underlying transition probability  depend on both the leader's and the follower's action. Such leader-follower interactions appear in other applications as well such as in AI Economist \cite{zheng2020ai},  Mechanism Design \cite{conitzer2004self}, optimal auction \cite{cole2014sample}, and security games \cite{tambe2011security}. 

Such kind of leader-follower interaction is different from the simultaneous play in the Markov setting as  considered in \cite{jin2021v,tian2021online}. In general, the Stackelberg equilibrium is the relevant concept for this type of leader-follower interaction \cite{conitzer2006computing} compared to the Nash equilibrium considered in the above paper.  \cite{letchford2009learning,peng2019learning} considered a learning framework in order to learn the Stackelberg equilibrium with a best-response oracle for the follower. However, these works can not be generalized to the {\em bandit feedback} setting where the leader and the follower are unaware of the transition probabilities and the rewards, and can only observe those rewards corresponding to the state-actions pairs encountered.  Efficient learning  in this leader-follower MDP setting under the bandit feedback (which is more natural) is fundamentally more challenging compared to the  single agent setting due to the more challenging exploration-exploitation trade-off. 

Few recent works have focused on such kind of leader-follower interaction in MDP with the bandit feedback setting. \cite{bai2021sample} considered  a model where the leader selects the underlying MDP on which the follower acts on, i.e., the leader only acts once at the start of the episode. However, we consider the setup where both the leader and the follower interact at every step of the episode.   \cite{kao2022decentralized} considers a setup where both the leader and the follower interact at every step similar to ours. However, \cite{kao2022decentralized} considered the setup where both the leader and the follower receive the same reward whereas in our setting the rewards can be different. Further, both the above papers consider the finite state-space (a.k.a. tabular setup) where the sample complexity scales with the state space. Thus, the above approach {\em would not be useful} for  large-scale RL applications where the number of states could even be infinite.  To address this curse of dimensionality, modern RL has adopted \emph{function approximation} techniques to approximate the (action-) value function or a policy, which greatly expands the potential reach of RL, especially via deep neural networks.  For large state-space the model-based approaches as considered in the above papers have limited application \cite{wei2020model}, thus, we focus on  developing model-free algorithm. Only \cite{zhong2021can} considers a leader-follower Markov setup with function approximation. However, they consider {\em myopic } followers which seek to maximize instantaneous reward and also consider the followers' reward are known.  Hence, it greatly alleviates the exploration challenge as it is only limited to  the leader's side. Rather, we consider the setup where the followers are also non-myopic with bandit feedback. Thus, we seek to answer the following question
\begin{center}
   \emph{Can we achieve provably optimal regret for model-free exploration for leader-follower (non-myopic) interaction in MDP with function approximation under bandit feedback?} 
\end{center}

\textbf{Our Contribution}: To answer the above question, we consider the Markov game with linear function approximation (bandit feedback) where at each step of the sequential decision process the leader takes an action and observing the action the follower reacts. The   transition probability and the reward functions can be represented as a {\em linear function} of some known feature mapping adapted from the single agent set up in \cite{jin2020provably}.  Our main contributions are:

\begin{itemize}[leftmargin=*]
\item We show that with a proper parameter choice, our proposed model-free algorithm achieves  $\tilde{\mathcal{O}}(\sqrt{d^3H^3T})$ regret for both leader and follower, where $d$ is the dimension of the feature mapping, $H$ is the length of the episode, and $T$ is the total number of steps.  Note that for the single agent setup, the regret is of the same order achieved in \cite{jin2020provably}. Hence, our result matches the regret bound for the single agent setup.

\item Our bounds are attained without explicitly estimating the unknown transition model or {\em requiring any simulator or best-response oracle}, and they depend on the state space only through the dimension of the feature mapping.  To the best of knowledge, these sub-linear regret bounds are the first results for  leader-follower non-myopic MDP game  with  function approximations under bandit feedback. 

\item Since linear MDP contains tabular setup, as a by-product, we provide the first result on the regret bound for the leader-follower (non myopic) MDP game under bandit feedback using model-free RL algorithm.
\item We adapt the classic model-free LSVI-UCB algorithm proposed in \cite{jin2020provably} in a novel manner. Due to the nature of the leader-follower interaction, a key challenge arises while establishing the value-aware uniform concentration, which lies at the heart of the performance analysis of model-free exploration. In particular, for a single agent set-up the greedy selection with respect to the standard $Q$-function achieves a small covering number. However, in the game setting, for a given policy, the best response strategy of a player fails to achieve such non-trivial covering number for the value-function class of the players   (i.e.,  $V$-function). To address this fundamental issue, we instead adopt a soft-max policy for the players by utilizing its nice property of approximation-smoothness trade-off via its parameter, i.e., temperature coefficient. 
\end{itemize}



\textbf{Related Literature}: Provably efficient RL algorithms for zero-sum Markov games have been proposed \cite{wei2017online,bai2020near,liu2021sharp,xie2020learning,chen2021almost,sayin2021decentralized}.  Provably efficient algorithms to obtain coarse correlated equilibrium have also been proposed for the general sum-game as well \cite{bai2020provable,jin2021v,mao2022provably}. In contrast to the above papers, we consider a leader-follower setup. Hence, our work is not directly comparable.

 For learning Stackelberg equilibrium, most of the works focus on the normal form game which is equivalent to step size $H=1$ in our setting \cite{balcan2015commitment,blum2014learning,peng2019learning,letchford2009learning}. Further, in the above papers, it is assumed that the followers' responses are known, in contrast, in our setting the follower is also learning its optimal policy.  \cite{zhong2021can} considered a leader-follower setup with myopic follower and known reward for linear MDP which alleviates the challenges of exploration for the follower.  \cite{zhong2021can} also proposed a model-based approach for tabular setup with myopic follower and unknown reward. In contrast, we consider linear MDP and non-myopic follower with bandit feedback, and proposed model-free RL algorithm. Recently, \cite{kao2022decentralized} proposed a decentralized cooperative RL algorithm with hierarchical information structure for a tabular set-up.  In contrast, we consider linear function approximation setup. In our model, the rewards of the leader and the follower can be different unlike in \cite{kao2022decentralized}.  \cite{bai2021sample} provides sample complexity guarantee for Stackelberg equilibrium for bandit-RL game where leader only takes action at the start of the episode which is quite different from our setup.  \cite{zheng2022stackelberg} modeled the actor-critic framework as a Stackelberg game which is quite different from our framework. 

\section{Leader-Follower MDP Game}
\textbf{The Model}: We consider an  episodic MDP with the tuple $(\mathcal{S},\mathcal{A},\cB,\mathcal{P},H,\mathbf{R})$ where $\mathcal{S}$ is the state space, $\mathcal{A}$ is the action space for leader, and $\mathcal{B}$ is the action space for followers.    Each MDP starts from the state $x_1$. At every step $h$, observing the state $x_h$, the leader first takes an action $a_h\in \cA$, then the follower takes an action $b_h\in \cB$ observing the action of the leader and the state $x_h$.  The state transitions to  $x_{h+1}\in \cS$ depending on $x_h$, $a_h,$ and $b_h$. The process continues for $H$ steps.  The transition probability kernel is defined as the following, $\mathcal{P}=\{\bP_h\}_{h=1}^{H}$ $\bP_h:\mathcal{S}\times\mathcal{A}\times \cB\rightarrow \mathcal{S}$.  The reward vector for leader and follower are defined as $\{r_{l,h}\}_{h=1}^{H}$ and $\{r_{f,h}\}_{h=1}^H$ respectively. The reward for agent $m=l,f$ $r_{m,h}(x_h,a_h,b_h)$ denotes the reward received by agent $m$ when the leader selects the action $a_h$ and the follower selects the action $b_{h}$ at step $h$. 


Several important points on the model should be noted here. This setup is also known as hierarchical MDP \cite{kao2022decentralized}. Our setup can model various real-world applications. For example, consider a dynamic electricity market where a social planner  sets a price at every hour of a day. Observing the price at current period, the users decide how much to consume at that period. Here, the state can represent the demand of the user, and the supply. The user's objective is to maximize its total utility over the day whereas the utility company's objective is  to maximize the social welfare. Note that such kind of leader-follower decision hierarchy can also occur in MARL when one agent has some advantage (e.g., can compute decision faster compared to the other agent) over the other agent.


Throughout this paper, we consider deterministic (unknown) reward. Without loss of generality, we also assume $|r_{m,h}|\leq 1$ for all $m$ and $h$. Our model can be easily extended to the setup where the rewards are random yet bounded. Our model can also be extended to the setup where the initial state of each episodic MDP is drawn from a distribution. 

Note that we consider the state $x$ as a joint state of both the leader and the follower. Our work can be extended to the setup where the leader and the follower's states are decoupled with the underlying assumption that the leader can observe the follower's state.

\textbf{Policy}: The agents interact repeatedly over $K$ episodes. The policy of the leader at step $h\in [H]$ at episode $k\in [K]$ is $\pi_{l,h}^k(a|x_h^k)$ that denotes the probability with which action $a\in \cA$ is taken at step $h$ at episode $k$ when the state is $x_h^k$.    The policy for the follower  at step $h\in[H]$ at episode $[K]$ is $\pi_{f,h}^{k}(b|x_h,a_h^k)$ that denotes the probability with action $b\in \cB$ is chosen by the follower at state $x_{h}$ at the $h$-th step of episode $k$ and when the  leader's action is $a_h^k$. Note the difference with the simultaneous play, here, the follower's policy  is a function of the leader's action at step $h$ whereas in the simultaneous game, it is independent of the other players' actions. Let $\pi_l=\{\pi_{l,h}\}_{h=1}^H$, and $\pi_f=\{\pi_{f,h}\}_{h=1}^H$ be the collection of the policies of leader and follower respectively across the episode. 


\textbf{Q-function and Value function}
The joint state-action value function for any player $m$, for $m=l,f$ at step $h$ is 
\begingroup\makeatletter\def\f@size{8.5}\check@mathfonts
\def\maketag@@@#1{\hbox{\m@th\large\normalfont#1}}
\begin{align}
    Q_{m,h}^{\pi_l,\pi_f}(x,a,b)=\mathbbm{E}\left[\sum_{i=h}^Hr_{m,i}(x_i,a_i,b_i)|x_h=x,a_h=a, b_h=b\right]\nonumber
\end{align}
\endgroup
Here, the expectation is taken over the transition probability kernel, and the policies of both the leader and the follower.  
We can also define a {\em marginal} $Q$-function for the leader as
\begin{align}\label{eq:q_fnmarginal}
q_{l,h}^{\pi_l,\pi_f}(x,a)=\sum_{b}\pi_{f,h}(b|x,a)Q_{l,h}^{\pi_l,\pi_f}(x,a,b).
\end{align}
The above denotes the expected cumulative reward  starting from step $h$ after playing action $a$, and then following the policy $\pi_l$ (from step $h+1$) while the follower following the policy $\pi_f$ from step $h$. We later show that marginal $q$ plays an important role in decision making.

    


For compactness of operator, we introduce the following notations
\begingroup\makeatletter\def\f@size{9}\check@mathfonts
\def\maketag@@@#1{\hbox{\m@th\large\normalfont#1}}
\begin{align}
    & \bD^{\pi_l,\pi_f}[Q](x)= \bE_{a\sim\pi_l(\cdot|x),b\sim\pi_f(\cdot|x,a)}Q(x,a,b),\nonumber\\ & \bD^{\pi_f}[Q](x,a)=\bE_{b\sim\pi_f(\cdot|x,a)}Q(x,a,b),\nonumber\\ & \bP_hV(x,a,b)=\bE_{x^{\prime}\sim\bP_h(\cdot|x,a,b)}V(x^{\prime}).\nonumber
\end{align}
\endgroup
The value function or expected cumulative reward  starting from step $h$ for the leader is defined as 
\begin{align}\label{eq:value_leader}
    V^{\pi_l,\pi_f}_{l,h}(x)=\bD^{\pi_{l,h},\pi_{f,h}}Q^{\pi_l,\pi_f}_{l,h}(x)
\end{align}
here the expectation is first taken over the follower's policy for a given leader's action then expectation is taken over the leader's action.



We now consider the follower's value functions. The leader's action dependent value function for the follower at step $h$ when the leader takes action $a$ at step $h$ is given by
\begin{align}\label{eq:q_fnmarginal_follower}
    \bar{V}_{f,h}^{\pi_l,\pi_f}(x,a)=\bD^{\pi_{f,h}}Q^{\pi_l,\pi_f}_{f,h}(x,a,b)
\end{align}
The above denotes the expected cumulative reward the follower would get from step $h$ after observing the leader's action at $h$, and then following its own policy at step $h$. 

We also define the following value function for the follower by taking the expectation over $a$ on $\bar{V}$
\begin{align}\label{eq:value_follower}
    V^{\pi_l,\pi_f}_{f,h}(x)=\bD^{\pi_{l,h},\pi_{f,h}}Q^{\pi_l,\pi_f}_{f,h}(x,a,b)
\end{align}
\textbf{Bellman's Equation}: We now describe the Bellman's equations for a given state and the joint action pair 
\begin{align}\label{bellman:follower}
    Q^{\pi_f}_{f,h}(x,a,b)=r_{f,h}(x,a,b)+\bP_hV_{f,h+1}^{\pi_l,\pi_f}(x,a,b)
\end{align}
Similarly, we have the following relationship
\begin{align}\label{bellman:leader}
    Q^{\pi_f}_{l,h}(x,a,b)=r_{l,h}(x,a,b)+\bP_hV_{l,h+1}^{\pi_l,\pi_f}(x,a,b)
\end{align}

\textbf{Information Structure}: We assume the following information structure--
\begin{assum}The transition probability kernel, and rewards are unknown to both the leader and the follower. The leader and follower observe the rewards  of each other only  for the encountered state-action pairs (i.e,{\em bandit feedback}). 
\end{assum}
Note that the agent can not access the rewards for the states and action which have not encountered before. This is also known as the {\em bandit-feedback} setting \cite{bai2021sample}. Thus, both the agents need to employ an exploratory policy.  Also, note that we consider an  information structure where the leader and follower can observe each other's reward.  Extending our analysis to the setup where the leader does not observe the action and/or reward of the follower constitutes a future research direction.   

\textbf{Objective}: After observing the action of the leader, the follower seeks to optimize its own leader's action dependent value function:
$
    \max_{\pi_f}\bar{V}^{\pi_l,\pi_f}_{f,1}(x_1,a)$

Given the policy of the follower, the leader seeks to optimize its own value function
\begin{equation*}
    \max_{\pi_l}V^{\pi_l,\pi_f}_{l,1}(x_1), \quad \pi_l^*=\arg \max_{\pi^l}V^{\pi_l,\pi_f}_{l,1}(x_1)
\end{equation*}

Now, we describe the relationship with $Q$-function and the value functions for the optimal policies of both the leader and the follower which will also specify how to select optimal policy at every step. For the follower, we have Bellman's optimality equation--
\begin{align}\label{eq:bellman_optimal}
    Q^{\pi_l,\pi_f^*}_{f,h}(x,a,b)=r_{f,h}(x,a,b)+ \bP_hV_{f,h+1}^{\pi_l,\pi_f^*}(x,a,b)
\end{align} 
Hence, given the action $a$ of the leader at step $h$, the follower's policy is 
$\pi_{f,h}^*(b|x,a)=1$ where $b=\max_{b}Q^{\pi_l,\pi_f}_{f,h}(x,a,b)$. Thus, the optimal policy is greedy with respect to the joint state-action $Q$ function. The follower's policy at step $h$ depends on the leader's action at step $h$, the leader's policy and the follower's own policy starting from $h+1$.  

The optimal policy $\pi_{l,h}^*$ for the leader if the follower selects the policy $\pi_f$ is given by
\begin{align}\label{eq:opt_leader}
    V^{\pi_l^*,\pi_f}_{l,h}(x_h)& =\max_{a}q^{\pi_l^*,\pi_f}_{l,h}(x_h,a)
\end{align}
Hence, the optimal policy for the leader is greedy with respect to its marginal $Q$-function $q_{l,h}^{\pi_l^*,\pi_f}(x_h,\cdot)$. The optimal policy at step $h$ depends on the follower's policy starting at step $h$, and the leader's policy starting from step $h+1$.

If the $Q$-functions are known, the optimal policy for  the leader and follower are obtained using backward induction. At step $H$, the follower's best response for every leader's action is computed ((as in (\ref{eq:bellman_optimal}). Then, leader's best response is computed  based on the marginal $q$ function (\ref{eq:opt_leader}). Once the policy is computed for the step $H$, the policy for step $H-1$ is computed in the similar manner and so on.




\textbf{Learning Metric}: Since the leader and the follower are unaware of the rewards and the transition probability, selecting optimal policy from the start is difficult. Rather, they seek to learn policies with good performance guarantee. Thus, instead of finding the Stackelberg equilibrium, we consider the following learning metric  
\begin{definition}
Regret for the leader is defined as
\begin{align}\label{regret:leader}
\mathrm{Regret}_{\ell}(K)=\sum_{k=1}^{K}(V_{l,1}^{\pi_l^{k,*},\pi_{f}^k}(x_1)-V_{1}^{\pi_{l}^k,\pi_f^k}(x_1))
\end{align}
where $\pi_l^{k,*}$ is the optimal policy for the leader when the follower plays the policy $\pi_{f}^k$.
\end{definition}
Note that the regret for the leader defines the optimality gap between the policies employed by the leader and the best response policy (in hindsight) of the leader given the follower's policy at an episode. Note that the follower's policy is unknown, rather, the leader needs to reason about the follower's policy at episode $k$. Also note that  the leader's regret measures the gap between the policy $\pi_l^k$ and  the best policy $\pi_l^{k,*}$ of the leader in response to the follower's policy at episode $k$, rather than the optimal policy of the follower. This is because 
the follower is also learning its optimal policy, hence,  the policy $\pi_f^k$ would not be optimal. Thus, the regret considers how good the leader is doing compared to the policy employed by the follower across the episodes. Obviously, at different episodes, the optimal policy of the leader may be different. 

\begin{definition}
Regret for the follower is defined as 
\begin{align}\label{regret:follower}
    \mathrm{Regret}_f=\sum_{k=1}^{K}(\bar{V}_{f,1}^{\pi_l,\pi_f^*}(x_1,a_1^k)-\bar{V}_{f,1}^{\pi_l,\pi_f^k}(x_1,a_1^k))
\end{align}
where $\pi_f^{*}$ denotes the optimal policy for the follower and  $a_1^k$ is the action of the leader at the first step of episode $k$. 
\end{definition}

 The regret for the follower captures how good the follower's policy is compared to the optimal policy for a given initial action and the policy of the leader.  Even though the follower knows the initial action of the leader, it is unaware of the leader's policy. Rather,  the follower also needs to reason about the leader's policy starting from step $2$.  Both the leader and the follower seek to minimize their respective regrets. Initial action specific regret is unique to the leader-follower setup.

Note that \cite{zhong2021can} considers the setup where the leader takes action at every step of the MDP whereas the followers are myopic. They also consider that the rewards are known, thus, one can compute the best response of the follower at every step. Thus, the regret for the follower does not arise there. \cite{kao2022decentralized} considered the reward is the same, thus, they consider the regret of the joint policy rather than the individual agent's regret. 

In general, achieving sub-linear regret in multi-agent RL setup is more challenging compared to the single agent setup since the underlying environment of an agent may change depending on the other agent's policy \cite{tian2021online,bai2021sample}. Nevertheless, we obtain sub-linear regret for both the leader and follower.  



\textbf{ Leader-follower Linear MDP}: We consider a linear MDP set-up in order to handle large state-space.
\begin{assum}\label{assum:linearmdp}
We consider a linear MDP with the known (to both the players) feature map $\phi: \mathcal{S}\times\mathcal{A}\times\mathcal{B}\rightarrow\R^{d}$, if for any h, there exists $d$ {\em unknown} signed measures $\mu_h=\{\mu^1_h,\ldots,\mu^d_h\}$ over $\mathcal{S}$  such that for any $(x,a,x^{\prime})\in \mathcal{S}\times \mathcal{A}_1\times \mathcal{B}$, 
\begin{align}
\bP_h(x^{\prime}|x,a,b)=\langle\phi(x,a,b),\mu_h(x^{\prime})\rangle\nonumber
\end{align}
and there exists vectors $\theta_{l,h}$, $\theta_{f,h}\in \R^d$ such that for any $(x,a,b)\in \mathcal{S}\times\mathcal{A}_1\times\mathcal{A}_2$,
\begin{align*}
r_{l,h}(x,a,b)=\langle\phi(x,a,b),\theta_{l,h}\rangle\quad r_{f,h}(x,a)=\langle\phi(x,a,b),\theta_{f,h}\rangle
\end{align*}
\end{assum}

Note that such a linear MDP setup is considered for single agent scenario \cite{jin2020provably,yang2019sample}. Examples of linear MDP includes tabular setup. 

For the leader-follower linear MDP setup, we have--
\begin{lem}
 $Q^{\pi_l,\pi_f}_{m,h}(x,a,b)=\langle \phi(x,a,b), w^{\pi_l,\pi_f}_{m,h}\rangle$ $\forall m$.
\end{lem}
Thus, the $Q$-functions of both the leader and follower  are linear in the feature space. We can thus search over $w$ in order to find the optimal $Q$-function. 
 \begin{algorithm}[h]
%
	\caption{Leader's Model Free RL Algorithm} 
	\label{algo:model_free}
	\begin{algorithmic}[1]
	%
			\STATE \textbf{Initialization: } $w_{l,h}=0$, $w_{f,h}=0$, $\alpha_f=\dfrac{\log(|\cB|)\sqrt{K}}{H}$,
			$\alpha_l=\dfrac{\log(|\cA|)\sqrt{K}}{H}$,$\beta=C_1dH\sqrt{\log(4(\log(|\cB||\cA|)+2\log(|\cB|)\log(|\cA|))dT/p)}$
			
			\FOR {episodes $k=1,\ldots, K$}
		\STATE Receive the initial state $x_1^k$.
		\FOR {step $h=H,H-1,\ldots, 1$}
		\STATE $\Lambda^k_{h}\leftarrow \sum_{\tau=1}^{k-1}\phi(x_h^{\tau},a^{\tau}_h,b_h^{\tau})\phi(x_h^{\tau},a_h^{\tau}, b_h^{\tau})^T+\lambda \bI$
\STATE $w^k_{m,h}\leftarrow (\Lambda^k_h)^{-1}[\sum_{\tau=1}^{k-1}\phi(x_h^{\tau},a^{\tau}_h,b_{h}^{\tau})[r_{m,h}(x_h^{\tau},a^{\tau},b_{h}^{\tau})+V^k_{m,h+1}(x_{h+1}^{\tau})]]$

\STATE $Q^k_{m,h}(\cdot,\cdot,\cdot)\leftarrow \min\{\langle w_{m,h}^k,\phi(\cdot,\cdot,\cdot)\rangle+\beta(\phi(\cdot,\cdot,\cdot)^T(\Lambda^k_{h})^{-1}\phi(\cdot,\cdot,\cdot))^{1/2},H\}$

\FOR{$a\in \cA$}
\STATE $\pi_{f,h}^k(b|\cdot,a)=\dfrac{\exp(\alpha_f(Q^{k}_{f,h}(\cdot,a,b)))}{\sum_{b_m}\exp(\alpha_f(Q^{k}_{f,h}(\cdot,a,b_m)))}$
\STATE $q^{k}_{l,h}(\cdot,\cdot)\leftarrow \langle \pi_{f,h}^k(\cdot|\cdot,\cdot),Q^k_{l,h}(\cdot,\cdot,\cdot)\rangle$
\STATE $\bar{V}^k_{f,h}(\cdot,\cdot)\leftarrow \langle \pi_{f,h}^k(\cdot|\cdot,\cdot),Q^k_{f,h}(\cdot,\cdot,\cdot)\rangle$
\ENDFOR
\STATE $\pi_{l,h}^k(a|\cdot)=\dfrac{\exp(\alpha_l(q^{k}_{l,h}(\cdot,a)))}{\sum_{a_m}\exp(\alpha_l(q^{k}_{l,h}(\cdot,a_m)))}$

\STATE $V^k_{l,h}(\cdot)=\langle\pi_{l,h}^k,q^k_{l,h}(\cdot,\cdot)\rangle$
\STATE $V^k_{f,h}(\cdot)=\langle \pi_{l,h}^k,\bar{V}^k_{f,h}(\cdot,\cdot)\rangle$
		\ENDFOR
		
		\FOR{ $h=1,\ldots, H$}
		\FOR{$a\in \cA$}
		\STATE Compute  $Q_{f,h}^k(x_h^k,a,b)$, $Q_{l,h}^k(x_h^k,a,b)$ for all $b$.
		\STATE Compute policy $\pi_{f,h}^k(b|x_h^k,a)$ according to the Soft-max for $Q_{f,h}^k(x_h^k,a,\cdot)$ with parameter $\alpha_f$.
		\STATE  $q_{l,h}(x_1^k,a)=\sum_{b}\pi_{f,h}^k(b|x_h^k,a)Q_{l,h}^k(x_h^k,a,b)$
		\ENDFOR
		\STATE The leader takes action $a_h^k$ according to Soft-max policy with respect to $q^k_{l,h}(x_h^k,\cdot)$ with  parameter $\alpha_l$.
		\STATE The follower takes an action $b_h^k$ (Algorithm~\ref{algo:model_free_follower}) and observe $x_{h+1}^k\sim \bP_h(x_h^k,a_h^k,b_h^k)$
		\ENDFOR
\ENDFOR
	
	\end{algorithmic}
\end{algorithm}
\begin{algorithm}[h]
%
	\caption{Follower's Model Free RL Algorithm} 
	\label{algo:model_free_follower}
	\begin{algorithmic}[1]
	%
			
		\STATE Execute steps 1-14 of Algorithm~\ref{algo:model_free}.
		

		\FOR {step $h=1,\ldots,H$}
		\STATE Observe the action $a_h^k$ of the leader.
		\STATE Compute $Q_{f,h}^k(x_h^k,a_h^k,b)$,  $\pi_{f,h}^k(b|x_{h}^k,a^k)$ for all $b$ based on $w_{f,h}^k$.
\STATE The follower takes action $b_{h}^k\sim \pi_{f,h}^k(\cdot|x_h^k,a_h^k)$ and observe $x_{h+1}^k$. 
\ENDFOR
	
	\end{algorithmic}
\end{algorithm}
\vspace{-0.1in}
\section{Proposed Algorithm}

We now describe our proposed algorithms for the leader  (Algorithm~\ref{algo:model_free}) and the follower (Algorithm~\ref{algo:model_free_follower}). The algorithm   is based on  the LSVI-UCB \cite{jin2020provably} with some subtle differences which we will point out along our description.

 We first describe the leader's algorithm. Note that in order to obtain its  policy, the leader also needs to reason the policy the follower would play. Hence, the leader's algorithm also consists of how the follower selects its policy. The first part (steps 5-6) consists of updating the parameters $\Lambda^k_h, w^k_{l,h},w^k_{f,h}$ which are used to update the joint state-action value functions $Q^k_{m,h}$   and value functions $V^k_{m,h}$ for $m=l,f$. Note that the Steps 7-14 are not evaluated for each state, rather, they are evaluated only for the encountered states till episode $k-1$. Hence, we do not need to iterate over potentially infinite number of states.  $V_{m,H+1}^k=0$ for all $k$.

We now discuss the rationale behind updating $w_{m,h}^k$. We seek to obtain $w$ such that it approximates the $Q$-function since the $Q$-function is inner product of $w$ and $\phi$ (Lemma 1). Thus, we parameterize  $Q_{m,h}^{\pi_l,\pi_f}(\cdot,\cdot,\cdot)$ by a linear form $\langle w_{m,h}^k,\phi(\cdot,\cdot,\cdot)\rangle$. The intuition is to obtain $w_{m,h}^k$ from the Bellman's equation using the regularized least-square regression. However, there are challenges. We do not know $\bP_h$ in Bellman's equations ((\ref{bellman:follower}),  and (\ref{bellman:leader})) rather $\bP_hV^{\pi_l,\pi_{f}}_{m,h+1}$ should be replaced by the empirical samples.    We obtain $w_{m,h}^k$ for $m=l,f$ by solving the following regularized least-square problem
\vspace{-0.1in}
  \begingroup\makeatletter\def\f@size{9}\check@mathfonts
\def\maketag@@@#1{\hbox{\m@th\large\normalfont#1}}  
\begin{align}\label{eq:ls}
    w_{m,h}^k\leftarrow \arg\min_{w\in \R^{d}}&  \sum_{\tau=1}^{k-1}[r_{m,h}(x_h^{\tau},a_h^{\tau},b_h^{\tau})+V_{m,h+1}^k(x_{h+1}^{\tau}) \nonumber\\
     & -w^T\phi(x_h^{\tau},a_h^{\tau})]^2 +\lambda ||w||_2^2
\end{align}
\endgroup
where $V_{m,h+1}^k$ is the estimate of the value function $V_{m,h+1}$.  After we obtain $w_{m,h}^k$, we add an additional bonus term $\beta(\phi(\cdot,\cdot,\cdot)^T(\Lambda^k_h)^{-1}\phi(\cdot,\cdot,\cdot))^{1/2}$ similar to \cite{jin2020provably} to obtain $Q_{m,h}^k$. $\beta$ is constant which we will characterize in the next section. $\Lambda^k_h$ is the Gram matrix for the regularized least square problem. Such an additional term is used for upper confidence bound in LSVI-UCB \cite{jin2020provably} as well. The same additional term is used for both $Q^k_{l,h}$ and $Q^k_{f,h}$. This bonus term would ensure the exploration for both the leader and the follower.

Now, we describe how we estimate the value function $V_{m,h+1}^k$ function which we use in (\ref{eq:ls}). In order to update the value function, we need to compute the policy for the follower and the leader (cf.(\ref{eq:value_leader})).   Unlike   LSVI-UCB, we use the soft-max policy for both the leader and follower. Soft-max policy $\textsc{Soft-Max}_{\alpha}(\vX)=\{\textsc{Soft-Max}^i_{\alpha}(\vX)\}_{i=1}^{|\cL|}$ for any vector $\vX\in \mathbbm{R}^{|\cL|}$ is a  vector with the same dimension as in $\vX$ with parameter $\alpha$ where the $i$-th component
\vspace{-0.05in}
 \begingroup\makeatletter\def\f@size{9}\check@mathfonts
\begin{align}\label{eq:soft-max}
    \textsc{Soft-Max}^i_{\alpha}(\vX)=\dfrac{\exp(\alpha X_i)}{\sum_{n=1}^{|\cL|}\exp(\alpha X_n)}
\end{align}
\endgroup
In order to estimate value function, first, one needs to compute follower's policy at a given leader's action, and subsequently, the leader's policy needs to computed (cf.(\ref{eq:value_leader}) \& (\ref{eq:value_follower})). At step $h$, for every leader's action $a$, $\pi_{f,h,k}(b|x_h^{\tau},a)$ is computed based on the soft-max policy on the estimated $Q$-function for the follower $Q^k_{f,h}(x_h^{\tau},a,b)$ at step 9.  Based on the follower's policy, the leader updates its marginal $Q$ function $q_l$ at step 10 (cf.(\ref{eq:q_fnmarginal})) and the follower's leader's action  dependent value function $\bar{V}_{f,h}^k$ (cf.(\ref{eq:q_fnmarginal_follower})) at step 11. Now, the leader computes its policy based on  $q_{l,h}^k(\cdot,\cdot)$. Finally, We update the leader's and follower's value function based on the leader's policy at steps 13 and 14. 
 
 Note that when $\alpha_l=\alpha_f=\infty$, the policy of the follower and leader become equal to the greedy policy. The greedy-policy is optimal for leader with respect to its marginal $q$-function (Eq.(\ref{eq:opt_leader})) and for the follower with respect to its joint $Q$-function (cf.(\ref{eq:bellman_optimal})). However, we can not obtain optimal regret for the leader and follower with the greedy policy unlike the single agent scenario. 
 

 The last part consists of execution of the policy.  In order to find its optimal policy,  the leader computes $Q_{f,h}^k$ for each action of the leader $a$ (Line 17),  based on the already computed $w_{f,h}^k$. Once $Q_{f,h}^k$ is computed, the follower's policy is also computed based on the soft-max function (Step 18). Once the follower's policy is computed, the marginal $Q$-function for the leader $q_{l,h}^k(x_h^k,\cdot)$ is computed.   The leader then takes an action $a_h^k$ based on the soft-max policy with respect to $q_{l,1}^k$. The follower takes an action $b_h^k$ by observing the action $a_h^k$ which we will describe next.

As mentioned before, the steps of  the follower (Algorithm~\ref{algo:model_free_follower}) are already contained in the leader's algorithm. The follower also obtains its $w$ by solving (\ref{eq:ls}). Hence, the leader and the follower have the same updates on $w$.  The only difference is the execution as the follower executes its action based on the action taken by the leader at every step.  At step 4 of Algorithm~\ref{algo:model_free_follower}, the follower computes $Q$-function based on the current state $x_h^k$ and action $a_h^k$ of the leader.  The follower then chooses its action based on the soft-max policy on the $Q$-function.

The space and time complexities of Algorithms ~\ref{algo:model_free} and \ref{algo:model_free_follower} are of the same order as the LSVI-UCB. To be precise, the space complexity is $\cO(d^2H+d|\cA||\cB|T)$. When we compute  $(\Lambda_h^k)^{-1}$ using Sherman-Morrison formula, the computation of $V_{m,h+1}^k$ is dominated by computing $Q_{m,h+1}^k$ and the policy $\pi_{l}^k$, and $\pi_f^k$. Hence, it takes $\cO(d^2|\cA||\cB| T)$ time.
\section{Analysis}

\subsection{Main Results}
\begin{theorem}\label{thm:episodic}
Fix $p>0$. If we set  $\beta=C_1dH\sqrt{\iota}$ in Algorithm~\ref{algo:model_free} and Algorithm~\ref{algo:model_free_follower} where $\iota=\log((\log(|\cA||\cB|)+2\log(|\cA|)\log(|\cB|))4dT/p)$ for some absolute constant $C_1$, then with probability $(1-p)$, 
\begin{align}
& \mathrm{Regret}_l(K)\leq C\sqrt{d^3H^3T\iota^2}\nonumber\\
& \mathrm{Regret}_f(K)\leq C^{\prime}\sqrt{d^3H^3T\iota^2}\nonumber
\end{align}
where $T=KH$ for some absolute constants $C$, and $C^{\prime}$.
\end{theorem}

The result shows that regret for both the leader and the follower scale with $\tilde{\cO}(\sqrt{d^3H^3T})$. Note that for the single-agent scenario \cite{jin2020provably}, the order of the regret is the same. However, compared to \cite{jin2020provably}, there is an additional multiplicative $\log(|\cA|)$ and $\log(|\cB|)$  factor in the value of $\iota$ which arises because we use soft-max policy for the leader and the follower instead of the greedy policy. The regret bounds do not depend on the dimension of the state space, rather, it depends on the dimension of the feature space $d$. If there is no follower, we set $|\cB|=1$, and can achieve single agent's regret for the soft-max policy as well. {\em To the best of our knowledge this is the first result which shows $\tilde{\cO}(\sqrt{T})$ regret for both the leader and the follower in the model-free with function approximation. }

 We assume that the leader observes the follower's action (this is known as informed game \cite{tian2021online}). Under uninformed setting (where a player may not observe the action of other player), it is statistically hard to obtain sub-linear regret even in zero-sum game \cite{tian2021online}. Thus, \cite{tian2021online} shows a sub-linear regret under a weaker notion of regret. It remains to be seen under such information structure, whether such result holds in the leader-follower setup. 

\subsection{Outline of the Proof}\label{sec:outline}
In this section, we provide an outline of our proof. First, we analyze the regret-bound for the leaders. 

\textbf{Regret for Leader}: We decompose the regret term for the leader as the following
\begin{align}
    \sum_{k}\underbrace{(V^{\pi_l^{k,*},\pi_{f}^k}(x_1)-V^{k}_{l,1}(x_1))}_{\cT_1}+\underbrace{(V^{k}_{l,1}(x_1)-V^{\pi_{l}^k,\pi_{f}^k}(x_1))}_{\cT_2}\nonumber
\end{align}
$\cT_1$ denotes the optimism term, and $\cT_2$ denotes the model-prediction error.

\textbf{Uniform Concentration}: In order to bound $\cT_1$ and $\cT_2$, we need to bound the difference between the  estimated value function $V_{l,h}^k$ and the  value function $V_{l,h}^{\pi_l,\pi_{f}}$  for any joint policy $\pi_l$ and $\pi_f$. As in the single agent case, the key step is to control the  fluctuations in least-squares value iteration. In particular, we need to show that for all $(k,h)\in [K]\times [H]$ with high  probability 
\begingroup\makeatletter\def\f@size{8}\check@mathfonts
\def\maketag@@@#1{\hbox{\m@th\large\normalfont#1}}
\begin{equation*}
\norm{\sum_{\tau=1}^{k-1}\phi(x_h^{\tau},a_h^{\tau},b_h^{\tau}) \left[V^k_{l,h+1}(x_{h+1}^{\tau})-\bP_hV^k_{l,h+1}(x_{h}^{\tau},a_h^{\tau},b_h^{\tau})\right]}_{(\Lambda^k_h)^{-1}}
\end{equation*}
\endgroup
 is upper bounded by  $\cO(d\sqrt{\log K})$. To this end, value-aware uniform concentration is required to handle the dependence between $V_{l,h+1}^k$ and samples $\{x_{h+1}^{\tau}\}_{\tau=1}^{k-1}$, which renders the standard self-normalized inequality infeasible in the model-free setting. The general idea here  is to fix a function class $\cV_{l,h}$ in advance and then show that each possible value function in our algorithm $V_{l,h}^k$ is within this class and has  log $\epsilon$-covering number that only scales with $\cO(\log(K))$. 
\emph{In the following, we fix an  $h \in [H]$ and drop the subscript $h$ for notation simplicity.}

\textbf{Why Soft-max?}: We first note that this uniform concentration bound is the main motivation for us to choose a soft-max policy for the leader and the follower as we will see that the standard greedy policy would fail in this case. That is, in order to guarantee that for each possible $V_l^k$, there is an $\epsilon$-close function in $\mathcal{V}_l$, it would basically lead to a very large covering number. Soft-max resolves the issue. 

We first define the following class for $Q_m$-function for $m = l,f$.
 $\cQ_m=\{Q_m|Q_m(\cdot,\cdot,\cdot)=
\min\{\langle w_{m},\phi(\cdot,\cdot,\cdot)\rangle
 +\beta\sqrt{\phi(\cdot,\cdot,\cdot)^T(\Lambda^k_h)^{-1}\phi(\cdot,\cdot,\cdot)},H\}.$ 
Then, we define the class of marginal $q$ function of the leader $\hat{\cQ}_l$ for a given leader's action $a$.
   $ \hat{\cQ}_l(a) = \{q_l| q_l(\cdot,a) = \sum_b \pi(b|\cdot,a) Q_l(\cdot,a,b); Q_l \in \cQ_l, \pi \in \Pi\},$
where $\Pi$ is given by the following class 
 $  \Pi =\{\pi|\pi(b|\cdot,a) = \textsc{Soft-Max}^b_{\alpha_f}((Q_{f}(\cdot,a,\cdot));
  \forall b\in \cB, Q_f \in \cQ_f \},$
where $\textsc{Soft-Max}$ is defined in (\ref{eq:soft-max}).  The necessity of the above function class is that the value-function class depends on $q_l$, i.e.,
the class of the value function for leader is
$\cV_l=\{V_l|V_l(\cdot)=\sum_{a}\pi_l(a|\cdot)q_l(\cdot,a),q_l\in \hat{Q}_l(a), \forall a\in \cA\}$ where $\pi_l$ is again Soft-max policy with parameter $\alpha_l$.
In order to  compute $\epsilon$-covering for the class $\cV_l$ we need to compute $\epsilon$-covering for the class $\hat{\cQ}_l$.
At this moment, we can explain why the introduction of soft-max in our algorithm is critical. Suppose we follow the standard greedy selection for the follower, which corresponds to $\alpha_f = \infty$ in above. The key issue in this approach is that one needs a  large  $\epsilon$-covering for $\hat{\cQ}_l$ so that each possible $q_l^k$ can be well-approximated (i.e., $\epsilon$-close) by function in $\hat{\cQ}_l$. This is in sharp contrast to the single-agent case where $\cV_l$ has a  covering number polynomial in $K$. To see this difference, in the single agent case, we do not have the policy of the follower. Since the value function for the single agent only depends on the agent's own policy, by the fact that $\max$ is a contraction map, an $\epsilon$-covering of  $Q_l^k$  implies an $\epsilon$-covering of $V_l^k$ and meanwhile the $\epsilon-$ covering number of $\cQ_m$ is polynomial in $K$  (Lemma~\ref{lm:qcovering}) by standard arguments. This no longer holds in the leader-follower setup as the value function  now inherently depends  on the follower's policy. In particular, note that if the policy is greedy for the follower then an $\epsilon$-covering of $Q_l^k$ {\em fails} to be an $\epsilon$-covering of $q_l^k$ in general since the greedy policy is not smooth in that a slight change of the follower's $Q$-function could lead to a substantial change of the follower's policy  eventually the leader's marginal $q$-function $q_{l}^k$. This leads to a large  distance for \emph{leader's} value functions due to the different action choices for the follower, even though the $Q$-function for the follower are close (See Example~\ref{eg:greedy_fails} for an example). Hence, one can not approximate the leader's value function within $\epsilon$-bound using greedy policy based on the follower's $Q$-function.
This fact motivates us to turn to $\textsc{Soft-Max}_{\alpha}$, which is Lipschitz continuous with a Lipschitz constant at most $2\alpha$. Thus, our main idea is as follows. Given $Q_f^k$,  we can first find fixed $\tilde{Q}_f \in \cQ_f$ ,  such that $\norm{Q_f^k -\tilde{Q}_f}_{\infty} \le \epsilon_1$,  then, thanks to the smoothness of soft-max function, we have $\norm{\pi_k - \tilde{\pi}}_1 \le 2\alpha \epsilon_1$ (Lemma~\ref{lem:pi}) where $\tilde{\pi}$ is the soft-max policy based on $\tilde{Q}_f$. Hence, using the above we obtain $\sum_b\pi(b|x,a)Q^k_l(x,a,b)-\sum_b \tilde{\pi}(b|x,a)\tilde{Q}_l(x,a,b)\leq \epsilon^{\prime}$ where $\norm{Q_l^k -\tilde{Q}_l}_{\infty} \le \epsilon_2$ by carefully choosing $\epsilon_1$ and $\epsilon_2$ as the family of $Q$ functions have small $\epsilon$-covering number. Thus, one can show that $\hat{\cQ}_l(a)$ has a log $\epsilon$-covering number which scales as $\cO(\log(K))$ for every $a$ as well.

Once we show that $\hat{\cQ}_l(a)$ has log-$\epsilon$ covering number which scales with $\log(K)$, we can show that the log-$\epsilon$ covering number for $\cV_l$ also scales at most with $\log(K)$. In particular, we find $\tilde{q}_l$ for a given $q_l$ such that $||\tilde{q}_l-q_l||_{\infty}\leq \epsilon^{\prime}$. Now using the Lipschitz continuity of the soft-max policy for the leader, we then show that $||\pi_l-\tilde{\pi}_l||_{1}\leq 2\alpha_l\epsilon^{\prime}$. Hence, we can  find $\tilde{V}_l$ such that $||V_l-\tilde{V}_l||_{\infty}\leq \epsilon$ by carefully choosing $\epsilon^{\prime}$ which is enough to show that log $\epsilon$-covering number for the value function class scales at most with $\cO(\log(K))$. To summarize, the soft-max policies enable to obtain $\epsilon$-covering number for the value function class from the $\epsilon$-covering number for the joint state-action value classes $Q$. Using the fact that $\cQ_m$ has log $\epsilon$-covering number of the order $\log(K)$ we can also achieve the log $\epsilon$-covering number for the value function class as $\log(K)$.  

In fact, a larger value of $\alpha_f$ and $\alpha_l$ means that we need a smaller $\epsilon$, hence a larger covering number. 
Then, one may wonder if  we can choose an arbitrarily small value for $\alpha_f$ or $\alpha_l$. However, the leader's regret will be large if we choose too small $\alpha_l$ (similarly, if we choose too small $\alpha_f$ the regret for the follower would be large).  In particular, we obtain

\begin{lem}\label{lem:ucb_leader}
With probability $1-p/2$,
\begin{align}
    \cT_1\leq \dfrac{KH\log(|\cA|)}{\alpha_l}
\end{align}
For $\alpha_l=\dfrac{\log(|\cA|\sqrt{K}}{H}$, we have $\cT_1\leq \sqrt{K}H^2$
\end{lem}
 Thus, we can not set $\alpha_l$ too small. Note that unlike the single agent scenario, $\cT_1$ is not upper bounded by $0$ as the leader does not select policy based on the greedy action. 

Combining the Azuma-Hoeffding inequality with the above, we obtain
\begin{lem}\label{lem:conc}
With probability $1-p/2$, $\cT_2\leq \cO(\sqrt{d^3TH^3\iota^2})$.
\end{lem}





\textbf{Regret for Follower}: Now, we analyze the regret bound for the followers. We decompose the regret term similar to the leader in the following manner
\begingroup\makeatletter\def\f@size{8}\check@mathfonts
\def\maketag@@@#1{\hbox{\m@th\large\normalfont#1}}
\begin{align}
    \sum_{k}\underbrace{\bar{V}^{\pi^k_l,\pi^*_f}_{f,1}(x_1,a_1^k)-\bar{V}^{k}_{f,1}(x_1,a_1^k)}_{\cT_3}+\underbrace{\bar{V}^k_{f,1}(x_1,a_1^k)-\bar{V}^{\pi_l^k,\pi_{f}^k}_{f,1}(x_1,a_1^k)}_{\cT_4}\nonumber
\end{align}
\endgroup

The analysis again hinges on showing that the value function class $\cV_f$ has log $\epsilon$-covering number that is upper bounded by $\cO(\log(K))$.  Since the value function $V_f^k$  depends on the joint policy of the leader and the follower, one needs soft-max policy of the leader and the follower to show that $\epsilon$-covering number of the $Q$-function class is enough to bound $\epsilon$-covering number for $\cV_f$  by $\cO(\log(K))$. Similar to $\cT_1$ and $\cT_2$ we obtain

\begin{lem} \label{lem:follower}
With probability $1-p/2$
\begin{align}
    \cT_3\leq KH\dfrac{\log(|\cB|)}{\alpha_f},\quad \cT_4\leq \sqrt{d^3H^3T\iota^2}
\end{align}
 By choosing $\alpha_f=\dfrac{\log(|\cB|)\sqrt{K}}{H}$, we obtain $\cT_3\leq \sqrt{KH^4}$.
\end{lem}
As one can see one can not set too small value of $\alpha_f$ which would increase the regret. On the other hand, too large $\alpha_f$ would not enable us to obtain optimal regret bound as well.

\begin{remark}
We have assumed that the feature space $\phi$ is known. Note that feature space learning is an active area of research \cite{zhang2022efficient,agarwal2020flambe,modi2021model} for linear approximation setup. The most promising technique is to estimate the $Q$-function by jointly optimizing over $w$ and $\phi$. Neural networks can be used to obtain such $w$ and $\phi$. Using similar technique we can learn $\phi$, however, such a characterization is left for the future.
\end{remark}

\begin{remark}
Recently, regret analysis for zero-sum game setup has been extended to the non-linear MDP setup \cite{jin2021bellman,huang2021towards}. Extending our analysis towards non-linear MDP by combining their approach constitutes an interesting future research direction. \end{remark}

\section{Equilibrium Learning}
The Algorithms~\ref{algo:model_free} and \ref{algo:model_free_follower} also enables us to obtain equilibrium policies ($\epsilon$-close). 

\textbf{Finding Coarse Correlated Stackelberg Equilibrium}: We first define CCSE.
\begin{definition}
Joint policy $(\pi_l,\pi_f)$ is coarse-correlated Stackelberg equilibrium (CCSE) if for all $x$
\begin{align}
V^{\pi_l,\pi_f}_{l,1}(x)\geq V^{\pi^{\prime}_l,\pi_f}_{l,1}(x), \quad  V^{\pi_l,\pi_f}_{f,1}(x)\geq V^{\pi_l,\pi^{\prime}_f}_{f,1}(x)\nonumber
\end{align}
for any leader's policy $\pi^{\prime}_l$ and follower's policy $\pi^{\prime}_f$. The joint policy is $\epsilon$-CCSE if
   $ V^{\pi_l,\pi_f}_{l,1}(x)\geq V^{\pi^{\prime}_l,\pi_f}_{l,1}(x)-\epsilon, \quad V^{\pi_l,\pi_f}_{f,1}(x)\geq V^{\pi_l,\pi^{\prime}_f}_{f,1}(x)-\epsilon$.
\end{definition}
Now, we show that our algorithms indeed return a joint policy which is $\epsilon$-CCSE.

\begin{cor}\label{cor:ccse}
Consider the joint policy: the leader and follower jointly choose a $k\in [K]$ with prob. $1/K$, and then the leader and follower select the policy $\pi_l^k$ and $\pi_f^k$ respectively (returned by Algorithms~\ref{algo:model_free} and \ref{algo:model_free_follower}). Such a joint policy  is $\tilde{\cO}(\sqrt{d^3H^4/K})$-CCSE with probability $1-p$.
\end{cor}
The above result entails that in order to achieve $\epsilon$-CCSE one needs $\tilde{\cO}(1/\epsilon^2)$ episodes. This is the first such result for non-myopic leader and follower Markov game setup with function approximation. The agent only needs to coordinate on the random number to choose the episode index $k$. 

\textbf{Finding Stackelberg Equilibrium}: For a zero-sum game, CCSE coincides with the Stackelberg equilibrium. Hence, by Corollary 1, we also obtain $\epsilon$- Stackelberg equilibrium for zero-sum game.

For more general setting, we can combine the reward-free exploration proposed in \cite{wang2020reward,liu2021sharp} and the soft-max policy to obtain SE with self-play. In particular, we divide the total episodes in two-phases, exploration phase and exploitation phase. In exploration phase, both the leader and the follower only explore to reduce the confidence bound. Instead of true reward, the reward will be the bonus term, $||\phi(x,a,b)||_{(\Lambda_h^k)^{-1}}$ to both the leader and follower which will incentivize the players to explore similar to \cite{wang2020reward} (however, we have to use soft-max policy instead of greedy policy). In exploitation phase, the leader and follower obtain policy similar to Algorithm 1 and 2. The complete characterization is left for the future.

\section{Conclusion and Future Work}
We propose a model-free RL-based algorithm for both the leader and the follower. We have achieved $\mathcal{\tilde{O}}(\sqrt{d^3H^3T})$ regret for both the leader and the follower. We have extended the LSVI-UCB algorithm towards the leader-follower setup. We have underlined the technical challenges in doing so and explained how the greedy policy for the players fail to achieve an uniform concentration bound for individual value function. We show that a soft-max policy for the players can achieve the regret bound.

  Whether we can tighten this dependence on $d$ remains an important future research direction. Whether we can the tighten the dependence on $H$ also constitutes a future research direction. Finally, we consider one leader and one follower setup. Extending our setup to multiple leaders and followers constitutes an important future research direction. 

\clearpage
\newpage
\bibliography{ref}
\clearpage
\newpage
\appendix
\textbf{Notations}: Throughout the rest of this paper, we denote $V_{l,h}^k$, $V_{f,h}^k$, $Q_{l,h}^k, Q_{f,h}^k,w_{l,h}^k, w_{f,h}^k, \Lambda_h^k$ as the $Q$-value and the parameter values estimated at the episode $k$. We also denote the marginal $q$-function for leader as $q_{l,h}^k$ and the leader-action dependent estimated value function as $\bar{V}_{f,h}^k$ at episode $k$.  
To simplify the presentation, we denote $\phi_h^k=\phi(x_h^k,a_h^k,b_h^k)$. To simplify notation, we also sometimes denote $\pi=\{\pi_l,\pi_f\}$.

Without loss of generality, we assume $||\phi(x,a)||_2\leq 1$ for all $(x,a)\in \cS\times\cA$, $||\mu_h(\cS)||_2\leq \sqrt{d}$, $||\theta_{m,h}||_2\leq \sqrt{d}$ for $m=l,f$ and all $h\in [H]$.

\section{Preliminary Results}\label{sec:preliminary}
\begin{lem}
Under Assumption~\ref{assum:linearmdp}, for any fixed policy $\pi_f$, $\pi_l$, let $w_{m,h}^{\pi}$ be the corresponding weights such that $Q^{\pi}_{m,h}(x,a,b)=\langle \phi(x,a,b),w_{m,h}^{\pi}\rangle$, for $m\in \{l,f\}$, then we have for all $h\in [H]$, 
\begin{align}
||w_{m,h}^{\pi}||\leq 2H\sqrt{d}
\end{align}
\end{lem}
\begin{proof}
From the linearity of the action-value function, we have
\begin{align}\label{eq:linearinpolicy}
     Q_{m,h}^{\pi}(x,a,b)&=r_{m,h}(x,a,b)+\bP_hV_{m,h}^{\pi}(x,a,b)\nonumber\\
   &  = \langle \phi(x,a,b),\theta_{m,h}\rangle+\int_{\mathcal{S}}V_{m,h+1}^{\pi}(x^{\prime},a)\langle \phi(x,a,b),d\mu_h(x^{\prime})\rangle\nonumber\\
   &  =\langle\phi(x,a,b),w_{m,h}^{\pi}\rangle
\end{align}
where $w_{m,h}^{\pi}=\theta_{m,h}+\int_{\mathcal{S}}V_{m,h+1}^{\pi}(x^{\prime},a)d\mu_h(x^{\prime})$.

Now, $||\theta_{m,h}||\leq \sqrt{d}$, and $||\int_{\mathcal{S}}V_{j,h+1}^{\pi}(x^{\prime})d\mu_h(x^{\prime})||\leq H\sqrt{d}$. Thus, the result follows from (\ref{eq:linearinpolicy}). 
\end{proof}

\begin{lem}\label{lem:w}
For any $(k,h)$, the weight $w_{j,h}^k$ satisfies 
\begin{align}
||w_{m,h}^k||\leq 2H\sqrt{dk/\lambda}
\end{align}
\end{lem}
\begin{proof}
For any vector $v\in \mathcal{R}^d$ we have from the definition of $w_{m,h}^k$ and $V^k_{m,h+1}$ as
\begin{align}\label{eq:inter}
    |v^Tw_{m,h}^k|=|v^T(\Lambda_h^k)^{-1}\sum_{\tau=1}^{k-1}\phi^{\tau}_h(x_h^{\tau},a^{\tau},b_h^{\tau})(m_h(x_h^{\tau},a^{\tau},b_h^{\tau})+\sum_{a}\pi^k_{l,h+1}(a|x_{h+1}^{\tau})\sum_{b}\pi^k_{f,h+1}(b|x_{h+1}^{\tau},a)Q_{m,h+1}^k(x_{h+1}^{\tau},a,b))|
\end{align}
here $\pi_{l,h+1}^k$, and $\pi_{f,h+1}^k(\cdot|x,a)$ are the Soft-max policies with respect to the marginal $q$-function and the joint state-action value function.

Note that $Q_{m,h+1}^k(x,a,b)\leq H$ for any $(x,a,b)$. Hence, from (\ref{eq:inter}) we have
\begin{align}
    |v^Tw_{j,h}^k|& \leq  \sum_{\tau=1}^{k-1}|v^T(\Lambda_h^k)^{-1}\phi^{\tau}_h|.2H\nonumber\\
    & \leq \sqrt{\sum_{\tau=1}^{k-1}v^T(\Lambda^h_k)^{-1}v}\sqrt{\sum_{\tau=1}^{k-1}\phi_h^{\tau}(\Lambda_h^k)^{-1}\phi_h^{\tau}}.2H\nonumber\\
    & \leq 2H||v||\dfrac{\sqrt{dk}}{\sqrt{\lambda}}
\end{align}
Note that $||w_{m,h}^k||=\max_{v:||v||=1}|v^Tw_{m,h}^k|$. Hence, the result follows. 
\end{proof}

\section{Regret for Leader}
Let us recall the decomposition of the regret for the leader.
\begin{align}
    \mathrm{Regret}(K)=\sum_{k}\underbrace{(V^{\pi_l^{k,*},\pi_{f}^k}(x_1)-V^{k}_{l,1}(x_1))}_{\cT_1}+\underbrace{(V^{k}_{l,1}(x_1)-V^{\pi_{l}^k,\pi_{f}^k}(x_1))}_{\cT_2}
\end{align}
We now bound $\cT_1$ and $\cT_2$.

\subsection{Proof Outline}
Here, we provide a outline of the Proof. The common step in obtaining the bound for $\cT_1$ and $\cT_2$ is to bound the gap between $Q^{\pi_l,\pi_f}_{l,h}$ and $Q^{k}_{l,h}$ for every $h$ and $k$. In particular, we show that the gap between $\langle \phi(x,a,b), w_{l,h}^k\rangle$ and $\langle \phi(x,a,b), w_{l,h}^{\pi_l,\pi_f}(x,a,b)\rangle$ is upper bounded by expected gap between $V_{l,h+1}^k$ and $V_{l,h+1}^{\pi_l,\pi_f}$ plus a term (which is exactly equal to the bonus term) with high probability (Lemma~\ref{lem:leader_q_diff}). In order to prove Lemma~\ref{lem:leader_q_diff}, we state and prove Lemma~\ref{lem:phi} which lies in the heart of the proof.  Further, we show that if one use soft-max, the sub-optimality gap can be bounded (Lemma~\ref{lem:close_optimal}). Now, recall that we add the bonus term to $\langle \phi(x,a,b), w_{l,h}^k\rangle$ to obtain $Q_{l,h}^k$. Thus, combining all of the above we bound $\cT_1$ using backward induction. Using Lemma~\ref{lem:leader_q_diff}, we show that the difference between $V_{l,h}^k-V_{l,h}^{\pi_l,\pi_f}$ can be upper bounded by sum of Martingale differences and sum of the bonus terms (Lemma~\ref{lm:recursion_leader}). Thus, $\cT_2$ can be bounded using Azuma-Hoeffding inequality and elliptical potential lemma.

\subsection{Base Results}
First, we state and prove Lemma~\ref{lem:phi}.
In order to bound $\cT_1$ and $\cT_2$, we state and prove some base results. 
\begin{lem}\label{lem:phi}
There exists a constant $C_2$ such that for any fixed $p\in (0,1)$, if we let $\mathcal{E}$ be the event that
\begin{align}
\norm{\sum_{\tau=1}^{k-1}\phi_{h}^{\tau}[V_{l,h+1}^{k}(x_{h+1}^{\tau})-\bP_hV_{l,h+1}^{k}(x_h^{\tau},a^{\tau}_h,b_h^{\tau})]}_{(\Lambda_h^k)^{-1}}\leq C_2dH\sqrt{\chi}
\end{align}
for all $j\in \{r,g\}$, $\chi=\log[4(C_1+1)(\log(|\cA||\cB|)+2\log(|\cA|)\log(|\cB|)) dT/p]$, for some constant $C_2$, then $\Pr(\mathcal{E})=1-p/2$. 
\end{lem}
The proof of Lemma~\ref{lem:phi} is technical and relegated later. An extra $\log(|\cA|)$ term appears compared to the single agent case \cite{jin2020provably} because the temp. coefficients of the soft-max policies ($\alpha_l,\alpha_f$) appears in the covering number. 

We now recursively bound the difference between the value function maintained in Algorithm~\ref{algo:model_free} (without the bonus term) and the value function $V_{l,h+1}^{\pi_l,\pi_f}$ for any follower's policy. We bound this using the expected difference at the next step plus an error term. This error term can be upper bounded by the bonus term with a high-probability.

\begin{lem}\label{lem:leader_q_diff}
There exists an absolute constant $\beta=C_1dH\sqrt{\iota}$, $\iota=\log((\log(|\cA||\cB|)+2\log(|\cA|)\log(|\cB|))4dT/p)$, and for any fixed policy $\pi_f$, on the event $\mathcal{E}$ defined in Lemma~\ref{lem:phi}, we have 
\begin{align}
\langle \phi(x,a,b),w_{l,h}^k\rangle-Q_{l,h}^{\pi_l,\pi_f}(x,a,b)=\bP_h(V_{l,h+1}^k-V^{\pi_l,\pi_f}_{l,h+1})(x,a,b)+\Delta_h^k(x,a,b)
\end{align}
for some $\Delta_h^k(x,a)$ that satisfies $|\Delta_h^k(x,a,b)|\leq \beta\sqrt{\phi(x,a,b)^T(\Lambda_h^k)^{-1}\phi(x,a,b)}$.
\end{lem}
\begin{proof}
Note that $Q_{l,h}^{\pi_l,\pi_f}(x,a,b)=\langle\phi(x,a,b),w_{l,h}^{\pi_l,\pi_f}\rangle=r_{l,h}(x,a,b)+\bP_hV_{l,h+1}^{\pi_l,\pi_f}(x,a,b)$. 


Hence, we have
\begin{align}\label{eq:diff}
& w_{l,h}^k-w_{l,h}^{\pi_l,\pi_f}=
 (\Lambda_h^k)^{-1}\sum_{\tau=1}^{k-1}\phi_h^{\tau}[r_{l,h}^{\tau}+V_{l,h+1}^k(x_{h+1}^{\tau})]
 -w_{l,h}^{\pi_l,\pi_f}\nonumber\\
& =-\lambda(\Lambda_h^k)^{-1}(w_{l,h}^{\pi_l,\pi_f})+ (\Lambda_h^k)^{-1}\sum_{\tau=1}^{k-1}\phi_h^{\tau}[V_{l,h+1}^k(x_{h+1}^{\tau})-\bP_hV_{l,h+1}^k(x_{h}^{\tau},a_h^{\tau},b_h^{\tau})]\nonumber\\
& +(\Lambda_h^k)^{-1}\sum_{\tau=1}^{k-1}\phi_h^{\tau}[\bP_hV_{l,h+1}^{k}(x_{h}^{\tau},a_h^{\tau},b_h^{\tau})-\bP_hV_{l,h+1}^{\pi}(x_h^{\tau},a_h^{\tau},b_h^{\tau})]
\end{align}

Now, we bound each term in the right hand side of expression in (\ref{eq:diff}). We call those terms as $\mathbf{q}_1$, $\mathbf{q}_2$, and $\mathbf{q}_3$ respectively. 

First, note that 
\begin{align}\label{eq:diffq1}
|\langle\phi(x,a,b),\mathbf{q}_1\rangle|& =|\lambda\langle\phi(x,a,b),( \Lambda_h^k)^{-1}(w_{l,h}^{\pi_l,\pi_f})\rangle|\nonumber\\
& \leq \sqrt{\lambda}||w_{l,h}^{\pi_l,\pi_f}||\sqrt{\phi(x,a,b)^T(\Lambda_h^k)^{-1}\phi(x,a,b)}
\end{align}
Second, from Lemma~\ref{lem:phi}, for the event in $\mathcal{E}$, we have
\begin{align}\label{eq:diffq2}
|\langle\phi(x,a,b),\mathbf{q}_2\rangle|\leq CdH\sqrt{\chi}\sqrt{\phi(x,a,b)^T(\Lambda_h^k)^{-1}\phi(x,a,b)}
\end{align}
where $\chi=\log[4(C_1+1)(\log(|\cA||\cB|)+2\log(|\cA|)\log(|\cB|)) dT/p]$.

Third,
\begin{align}\label{eq:diff2}
& \langle \phi(x,a,b),\mathbf{q}_3\rangle=\langle \phi(x,a,b),(\Lambda_h^k)^{-1}\sum_{\tau=1}^{k-1}\phi_h^{\tau}[\bP_h(V_{l,h+1}^{k}-V_{l,h+1}^{\pi_l,\pi_f})(x_h^{\tau},a_h^{\tau},b_h^{\tau})]\rangle\nonumber\\
& =\langle\phi(x,a,b),(\Lambda_h^k)^{-1}\sum_{\tau=1}^{k-1}\phi_h^{\tau}(\phi_h^{\tau})^T\int(V_{l,h+1}^k-V_{l,h+1}^{\pi_l,\pi_f})(x^{\prime})d\mu_h(x^{\prime})\rangle\nonumber\\
& =\langle\phi(x,a,b),\int(V_{l,h+1}^k-V_{l,h+1}^{\pi_l,\pi_f})(x^{\prime})d\mu_h(x^{\prime})\rangle -\langle\phi(x,a,b),\lambda(\Lambda_h^k)^{-1}\int(V_{l,h+1}^k-V_{l,h+1}^{\pi_l,\pi_f})(x^{\prime})d\mu_h(x^{\prime})\rangle
\end{align}
The second term  in (\ref{eq:diff2}) can be bounded as the following
\begin{align}\label{eq:diffq31}
|\langle\phi(x,a,b),\lambda(\Lambda_h^k)^{-1}\int(V_{m,h+1}^k-V_{m,h+1}^{\pi_l,\pi_f})(x^{\prime})d\mu_h(x^{\prime})\rangle|\leq 2H\sqrt{d\lambda}\sqrt{\phi(x,a,b)^T(\Lambda_h^k)^{-1}\phi(x,a,b)}
\end{align}
since $||\int(V_{r,h+1}^k-V_{r,h+1}^{\pi})(x^{\prime})d\mu_h(x^{\prime})||_2\leq 2H\sqrt{d}$ as $||\mu_h(\cS)||\leq \sqrt{d}$. The first term in (\ref{eq:diff2}) is equal to
\begin{align}\label{eq:diffq32}
\bP_h(V_{m,h+1}^k-V_{m,h+1}^{\pi_l,\pi_f})(x,a,b)
\end{align}
Note that $\langle \phi(x,a,b),w_{l,h}^k\rangle-Q_{l,h}^{\pi_l,\pi_f}(x,a,b)=\langle \phi(x,a,b),w_{l,h}^k-w_{l,h}^{\pi_l,\pi_f}\rangle=\langle \phi(x,a,b),\mathbf{q_1}+\mathbf{q_2}+\mathbf{q_3}\rangle$. 
Since $\lambda=1$, we have from (\ref{eq:diffq1}), (\ref{eq:diffq2}),(\ref{eq:diffq31}), and (\ref{eq:diffq32})
\begin{align}
    |\langle\phi(x,a,b),w_{l,h}^k\rangle-Q_{l,h}^{\pi_l,\pi_f}(x,a,b)-\bP_h(V_{l,h+1}^k-V_{l,h+1}^{\pi_l,\pi_f})(x,a,b)| \leq C_3dH\sqrt{\chi}\sqrt{\phi(x,a,b)^T(\Lambda_h^k)^{-1}\phi(x,a,b)}
\end{align}
for some constant $C_3$ which is independent of $C_1$. Finally, note that 
\begin{align}\label{eq:const_beta}
    C_3\sqrt{\chi}&=\sqrt{\log[4(C_1+1)(\log(|\cA||\cB|)+2\log(|\cA|)\log(|\cB|))]} dT/p]\nonumber\\
    &= C_3\sqrt{\iota+\log(C_1+1)}\nonumber\\
    & \leq C_1\sqrt{\iota}
\end{align}
\end{proof}
\subsection{Proof of Lemma~\ref{lem:ucb_leader}}
Using the above, given a follower's policy, we  bound the gap between the optimal value function and the estimated value function $V_{l,h}^k$ maintained by our algorithm (Lemma~\ref{lem:ucb_leader}). 

Before that we state and prove another result.
\begin{lem}\label{lem:close_optimal}
Then, $\tilde{V}_{l,h}^k(x)-V_{l,h}^{k}(x)\leq \dfrac{\log|\cA|}{\alpha_l}$
\end{lem}
where 
\begin{definition}\label{defn:barvhk}
 $\tilde{V}_{l,h}^k(\cdot)=\max_{a}[\sum_{b}\pi_{f,h}^k(b|\cdot,a)Q_{l,h}^k(\cdot,a,b)]=\max_a q_{l,h}^k(\cdot,a)$.
\end{definition}
$\bar{V}_{l,h}^k(\cdot)$ is the value function corresponds to the greedy-policy with respect to the marginal $Q$ function. 



\begin{proof}
Note that
\begin{align}
V_{l,h}^{k}(x)=\sum_{a}\pi_{l,h}(a|x)\sum_{b}\pi_{f,h}^k(b|x,a)Q_{l,h}^k(x,a,b)
=\sum_{a}\pi_{l,h}(a|x)q_{l,h}^k(x,a)
\end{align}
where
\begin{align}\label{eq:boltz}
    \pi_{h,k}(a|x)=\dfrac{\exp(\alpha_l[q_{l,h}^k(x,a)])}{\sum_{a}\exp(\alpha_l[q_{l,h}^k(x,a)])}
\end{align}
Denote $a_x=\arg\max_{a}q_{l,h}^k(x,a)$

Now, recall from Definition~\ref{defn:barvhk} that $\tilde{V}_{h}^k(x)=q_{l,h}^k(a_x)$. Then,
\begin{align}\label{eq:uppb1}
    & \tilde{V}_{l,h}^k(x)-V_{l,h}^{k}(x)=q_{l,h}^k(a_x) - \sum_{a}\pi_{l,h}^k(a|x)q_{l,h}^k(x,a)\nonumber\\
    & \leq \left(\dfrac{\log(\sum_{a}\exp(\alpha_l(q_{l,h}^k(x,a))))}{\alpha_l}\right)- \sum_{a}\pi_{l,h}^k(a|x)q_{l,h}^k(x,a)\nonumber\\
  & \leq \dfrac{\log(|\cA|)}{\alpha_l}
\end{align}
where the last inequality follows from Proposition 1 in \cite{pan2019reinforcement}.
\end{proof}

We are now ready to prove Lemma~\ref{lem:ucb_leader}.

\begin{proof}
We prove the lemma by Induction.


First, we prove for the step $H$. 

 Note that $Q_{l,H+1}^k=0=Q_{l,H+1}^{\pi}$.

Under the event in $\mathcal{E}$ as described in Lemma~\ref{lem:phi} and \ref{lem:leader_q_diff}, we have 
\begin{align}
& |\langle\phi(x,a,b),w_{l,H}^k\rangle-Q_{l,H}^{\pi}(x,a,b)| \leq  \beta\sqrt{\phi(x,a,b)^T(\Lambda_H^k)^{-1}\phi(x,a,b)}\nonumber
\end{align}
Hence, for any $(x,a,b)$,
\begin{align}
 Q_{l,H}^{\pi}(x,a,b)& \leq \min\{\langle\phi(x,a,b),w_{l,H}^k\rangle+\beta\sqrt{\phi(x,a,b)^T(\Lambda_H^k)^{-1}\phi(x,a,b)},H\}\nonumber\\& 
= Q_{l,H}^k(x,a,b)
\end{align}
Since $\pi^k_f$ derived by the leader and the played by the follower is the same, thus, for any $(x,a)$
\begin{align}
    q_{l,H}^{\pi}(x,a)& =\sum_{b}\pi^k_f(b|x,a)Q^{\pi}_{l,H}(x,a,b)\nonumber\\
   &  \leq \sum_{b}\pi^k_f(b|x,a)Q^{k}_{l,H}(x,a,b)= q_{l,H}^{k}(x,a)
\end{align}
Hence, from the definition of $\tilde{V}_h^k$,
\begin{align}
\tilde{V}_{H}^k(x)& =\max_{a}q_{l,H}^k(x,a,b)\geq \sum_{a}\pi_{l,h}^k(a|x)q_{l,h}^k(x,a) \nonumber\\
& =V_{l,H}^{\pi_l,\pi_f}(x)
\end{align}
for any policy $\{\pi_l,\pi_f\}$.  Thus, it also holds for $\{\pi^{k,*}_l,\pi_f\}$, the optimal policy. Hence, from Lemma~\ref{lem:close_optimal}, we have 
\begin{align}
    V_{l,H}^{\pi_l^{k,*},\pi_f}(x)-V_{l,H}^{k}(x)\leq \dfrac{\log(|\cA|)}{\alpha_l}\nonumber
\end{align}

Now, suppose that it is true till the step $h+1$ and consider the step $h$.

Since, it is true till step $h+1$, thus, for any policy $\pi$,
\begin{align}
\bP_h(V_{l,h+1}^{\pi_l^{k,*},\pi_f}-V_{l,h+1}^k)(x,a,b)\leq \dfrac{(H-h)\log(|\cA|)}{\alpha_l}
\end{align}

From Lemma~\ref{lem:leader_q_diff} and the above result, we have for any $(x,a,b)$
\begin{align}
& Q_{l,h}^{\pi_l^{k,*},\pi_f}(x,a,b)  \leq  Q_{l,h}^{k}(x,a,b)+\dfrac{(H-h)\log(|\cA|)}{\alpha_l}
\end{align}
Hence, for any $(x,a)$ and the fact that $\sum_{b}\pi_{f,h}(b|x,a)=1$, we have
\begin{align}
    q^{\pi_l^{k,*},\pi_f}_{l,h}(x,a)\leq q^{k}_{l,h}(x,a)+\dfrac{(H-h)\log(|\cA|)}{\alpha_l}
\end{align}
Thus, from the definition of $\tilde{V}_{l,h}^k$
\begin{align}
    V^{\pi_l^{k,*},\pi_f}_{l,h}(x)\leq \tilde{V}_{l,h}^k(x)+\dfrac{(H-h)\log(|\cA|)}{\alpha_l}
\end{align}
Now, again from Lemma~\ref{lem:close_optimal}, we have $\tilde{V}_{l,h}^{k}(x)-V_{l,h}^k(x)\leq \dfrac{\log(|\cA|)}{\alpha}$. Thus,
\begin{align}
    V^{\pi_l^{k,*},\pi_f}_{l,h}(x)-V_{l,h}^k(x)\leq \dfrac{(H-h+1)\log(|\cA|)}{\alpha_l}
\end{align}
Thus, we have
\begin{align}
    V^{\pi^*_l,\pi_f}_{l,h}(x)-V_{l,h}^k(x)\leq \dfrac{(H-h+1)\log(|\cA|)}{\alpha_l}\nonumber
\end{align}
Hence, the result follows by summing over $K$ and considering $h=1$. 
\end{proof}

In order to prove the Lemma~\ref{lem:conc}, we state and prove another result. 
 
 First, we introduce a notation. 
 Let
\begin{align}\label{eq:d_martingale_leader}
    & D_{l,h,1}^k=\bD^{\pi_l^k,\pi_f^k}(Q_{l,h}^k-Q_{l,h}^{\pi_l^k,\pi_f^k})-(Q_{l,h}^{k}(x_h^k,a_h^k,b_h^k)-Q_{l,h}^{\pi^k}(x_h^k,a_h^k,b_h^k))\nonumber\\
    & D_{l,h,2}^k=\bP_h(V_{l,h+1}^k-V_{j,h+1}^{\pi_k})(x_h^k,a_h^k,b_h^k)-[V_{l,h+1}^k-V_{l,h+1}^{\pi^k}](x_{h+1}^k)
\end{align}
We will show that the above two terns are Martingale under proper filtration which we use to prove Lemma~\ref{lem:conc}. First, we show that the difference between $V_{l,h}^k(x_1)-V_{l,h}^{\pi_l^k,\pi_f^k}$ as the sum of the above two martingales  plus the bonus term. In the following, we denote $\pi^k=\{\pi_l^k,\pi_f^k\}$.
\begin{lem}\label{lm:recursion_leader}
 On the event defined in $\mathcal{E}$ in Lemma~\ref{lem:phi}, we have
 \begin{align}
    V_{l,1}^{k}(x_1)-V_{l,1}^{\pi_l^k,\pi_f^k}(x_1)\leq\sum_{h=1}^{H}(D_{l,h,1}^k+D_{l,h,2}^k)+\sum_{h=1}^{H}2\beta\sqrt{\phi(x_{h}^k,a_{h}^k,b_h^k)^T(\Lambda_h^k)^{-1}\phi(x_{h}^k,a_{h}^k,b_h^k)}\nonumber
\end{align}
\end{lem}
\begin{proof}
By Lemma~\ref{lem:leader_q_diff}, for any $x,h,a,b,k$
\begin{align}
& \langle w_{l,h}^k,\phi(x,a,b)\rangle+\beta\sqrt{\phi(x,a,b)^T(\Lambda_h^k)^{-1}\phi(x,a,b)}-Q_{j,h}^{\pi^k}(x,a,b) \nonumber\\
& \leq \bP_h(V_{l,h+1}^k-V_{l,h+1}^{\pi^k})(x,a,b)+2\beta\sqrt{\phi(x,a,b)^T(\Lambda_h^k)^{-1}\phi(x,a,b)}
\end{align}
Thus, 
\begin{align}\label{eq:q_d_leader}
Q_{l,h}^{k}(x,a,b)-Q_{l,h}^{\pi^k}(x,a,b)\leq \bP_h(V_{l,h+1}^k-V_{l,h+1}^{\pi^k})(x,a,b)+2\beta\sqrt{\phi(x,a,b)^T(\Lambda_h^k)^{-1}\phi(x,a,b)}\nonumber\\
\bP_h(V_{l,h+1}^k-V_{l,h+1}^{\pi^k})(x,a,b)+2\beta\sqrt{\phi(x,a,b)^T(\Lambda_h^k)^{-1}\phi(x,a,b)}-(Q_{l,h}^{k}(x,a,b)-Q_{l,h}^{\pi^k}(x,a,b))\geq 0
\end{align}
Since $V_{l,h}^{k}(x)=\bD^{\pi_{l,h}^k,\pi_{f,h}^k}Q_{l,h}^k(x)$ and $V_{l,h}^{\pi^k}(x)=\bD^{\pi_{l,h}^k,\pi_{f,h}^k}Q_{l,h}^{\pi^k}(x)$. 

Thus, from (\ref{eq:q_d_leader}),
\begin{align}\label{eq:recursive_leader}
    & V_{l,h}^{k}(x_h^k)-V_{l,h}^{\pi_k}(x_h^k)=\sum_{a}\sum_{b}\pi_{l,h}^k(a|x_{h}^k)\pi_{f,h}^k(b|x_h^k,)[Q_{l,h}^{k}(x_{h}^k,a,b)-Q_{l,h}^{\pi^k}(x_{h}^k,a,b)]\nonumber\\
    & \leq \sum_{a}\sum_{b}\pi_{l,h}^k(a|x_h^k)\pi_{f,h}^k(b|x_h^k,a)[Q_{j,h}^{k}(x_{h}^k,a,b)-Q_{j,h}^{\pi^k}(x_{h}^k,a,b)]\nonumber\\
    & +2\beta\sqrt{\phi(x_{h}^k,a_{h}^k,b_h^k)^T(\Lambda_h^k)^{-1}\phi(x_{h}^k,a_{h}^k,b_h^k)}+\bP_h(V_{l,h+1}^k-V_{l,h+1}^{\pi^k})(x_{h}^k,a_h^k,b_h^k)-(Q_{l,h}^{k}(x_h^k,a_h^k,b_h^k)-Q_{l,h}^{\pi^k}(x_h^k,a_h^k))
\end{align}

Thus, from (\ref{eq:recursive_leader}), we have
\begin{align}
    V_{l,h}^k(x_h^k)-V_{l,h}^{\pi^k}(x_h^k)\leq D_{l,h,1}^k+D_{l,h,2}^k+[V_{l,h+1}^k-V_{l,h+1}^{\pi^k}](x_{h+1}^k)+2\beta\sqrt{\phi(x_{h}^k,a_{h}^k,b_h^k)^T(\Lambda_h^k)^{-1}\phi(x_{h}^k,a_{h}^k,b_h^k)}\nonumber
\end{align}
Hence, by iterating recursively, we have
\begin{align}
    V_{l,1}^{k}(x_1)-V_{l,1}^{\pi^k}(x_1)\leq\sum_{h=1}^{H}(D_{l,h,1}^k+D_{l,h,2}^k)+\sum_{h=1}^{H}2\beta\sqrt{\phi(x_{h}^k,a_{h}^k,b_h^k)^T(\Lambda_h^k)^{-1}\phi(x_{h}^k,a_{h}^k,b_h^k)}\nonumber
\end{align}
The result follows.
\end{proof}

\begin{proof}
Note from Lemma~\ref{lm:recursion_leader}, we have
\begin{align}\label{eq:martin}
\sum_{k=1}^{K}V_{l,1}^{k}(x_1)-V_{l,1}^{\pi_k}(x_1)\leq \sum_{k=1}^{K}\sum_{h=1}^{H}(D_{l,h,1}^k+D_{l,h,2}^k)+\sum_{k=1}^{K}\sum_{h=1}^{H}2\beta\sqrt{\phi(x_h^k,a_h^k,b_h^k)^T(\Lambda_h^k)^{-1}\phi(x_h^k,a_h^k,b_h^k)}
\end{align}
We, now, bound the individual terms. First, we show that the first term corresponds to a Martingale difference.

For any $(k,h)\in [K]\times [H]$, we define $\cF_{h,1}^k$ as $\sigma$-algebra generated by the joint state-action sequences,  $\{(x_i^{\tau},a_i^{\tau},b_i^{\tau})\}_{(\tau,i)\in [k-1]\times [H]}\cup \{(x^k_{i},a_i^k,b_i^k)\}_{i\in [h]}$. 

Similarly, we define the $\cF_{h,2}^k$ as the $\sigma$-algebra generated by $\{(x_i^{\tau},a_i^{\tau},b_i^{\tau})\}_{(\tau,i)\in [k-1]\times [H]}\cup \{(x^k_{i},a_i^k,b_i^k)\}_{i\in [h]}\cup\{x_{h+1}^k\}$. $x_{H+1}^k$ is a null state for any $k\in [K]$. 

A filtration is a sequence of $\sigma$-algebras $\{\cF_{h,m}^k\}_{(k,h,m)\in [K]\times[H]\times[2]}$ in terms of time index
\begin{align}
    t(k,h,m)=2(k-1)H+2(h-1)+m\nonumber
\end{align}
which holds that $\cF_{h,m}^k\subset \cF_{h^{\prime},m^{\prime}}^{k^{\prime}}$ for any $t\leq t^{\prime}$. 

Note from the definitions in (\ref{eq:d_martingale_leader}) that $D_{l,h,1}^k\in \mathcal{F}_{h,1}^k$ and $D_{l,h,2}^k\in \mathcal{F}_{h,2}^k$. Thus, for any $(k,h)\in [K]\times [H]$, 
\begin{align}
    \mathbbm{E}[D_{l,h,1}^k|\cF_{h-1,2}^k]=0, \quad \mathbbm{E}[D_{l,h,2}^k|\cF_{h,1}^k]=0\nonumber
\end{align}
Notice that $t(k,0,2)=t(k-1,H,2)=2(H-1)k$. Clearly, $\cF_{0,2}^k=\cF_{H,2}^{k-1}$ for any $k\geq 2$. Let $\cF_{0,2}^1$ be empty. We define a Martingale sequence
\begin{align}
    M_{l,h,m}^k& = \sum_{\tau=1}^{k-1}\sum_{i=1}^{H}(D_{l,i,1}^{\tau}+D_{l,i,2}^{\tau})+\sum_{i=1}^{h-1}(D_{l,i,1}^{k}+D_{l,i,2}^k)+\sum_{l^{\prime}=1}^{m}D_{l,h,l^{\prime}}^k\nonumber\\
    & =\sum_{(\tau,i,l^{\prime})\in [K]\times[H]\times[2], t(\tau,i,l^{\prime})\leq t(k,h,m)}D^{\tau}_{l,i,l^{\prime}}\nonumber
\end{align}
where $t(k,h,m)=2(k-1)H+2(h-1)+m$ is the time index. Clearly, this martingale is adopted to the filtration $\{\cF_{h,m}^k\}_{(k,h,m)\in [K]\times [H]\times [2]}$, and particularly
\begin{align}\label{eq:leader_sum_last_martingale}
    \sum_{k=1}^K\sum_{h=1}^H(D_{l,h,1}^k+D_{l,h,2}^k)=M_{l,H,2}^K
\end{align}

Thus, $M_{l,H,2}^K$ is a Martingale difference satisfying $|M_{l,H,2}^K|\leq 4H$ since $|D_{l,h,1}^k|,|D_{l,h,2}^k|\leq 2H$
From the Azuma-Hoeffding inequality, we have
\begin{align}
\Pr(M_{l,H,2}^K> s)\leq 2\exp(-\dfrac{s^2}{16TH^2})\nonumber
\end{align}
With probability $1-p/4$ at least
\begin{align}\label{eq:bound_last_martingale}
\sum_{k}\sum_{h}M_{l,H,2}^K\leq \sqrt{16TH^2\log(4/p)}
\end{align}

Now, we bound the second term in (\ref{eq:martin}). Note that the minimum eigen value of $\Lambda_h^k$ is at least $\lambda=1$ for all $(k,h)\in [K]\times [H]$. By Lemma~\ref{lem:le1}, 
\begin{align}
\sum_{k=1}^{K}(\phi_h^k)^T(\Lambda_h^k)^{-1}\phi_h^k\leq 2\log\left[\dfrac{\det(\Lambda_h^{k+1})}{\det(\Lambda_h^1)}\right]\nonumber
\end{align}
Moreover, note that $||\Lambda_h^{k+1}||=||\sum_{\tau=1}^{k}\phi_h^k(\phi_h^k)^T+\lambda\bI||\leq \lambda+k$, hence,
\begin{align}
   \sum_{k=1}^{K} (\phi_h^k)^T(\Lambda_h^k)^{-1}\phi_h^k\leq 2d\log\left[\dfrac{\lambda+k}{\lambda}\right]\leq 2d\iota\nonumber
\end{align}
Now, by Cauchy-Schwartz inequality, we have
\begin{align}\label{eq:bonus_leader_bound}
    \sum_{k=1}^{K}\sum_{h=1}^H \sqrt{(\phi_h^k)^T(\Lambda_h^k)^{-1}\phi_h^k}& \leq \sum_{h=1}^{H}\sqrt{K}[\sum_{k=1}^{K}(\phi_h^k)^T(\Lambda_h^k)^{-1}\phi_h^k]^{1/2}\nonumber\\
    & \leq H\sqrt{2dK\iota}
\end{align}
Note that $\beta=C_1dH\sqrt{\iota}$. 

Thus, combining (\ref{eq:leader_sum_last_martingale}), (\ref{eq:bound_last_martingale}), and (\ref{eq:bonus_leader_bound}), we have with probability $1-p/2$,
\begin{align}
\sum_{k=1}^{K}V_{l,1}^{k}(x_1^{k})-V_{l,1}^{\pi_k}(x_1^k)
\leq  [\sqrt{16TH^2\log(4/p)}+C_4\sqrt{d^3H^3T\iota^2}]\nonumber
\end{align}
Hence, the result follows. 
\end{proof}

\section{Regret for Follower}

We now prove the regret bound for the followers. Recall the decomposition of the regret term for follower.

\begin{align}
    \sum_{k}\underbrace{\bar{V}^{\pi^k_l,\pi^*_f}_{f,1}(x_1,a_1^k)-\bar{V}^{k}_{f,1}(x_1,a_1^k)}_{\cT_3}+\underbrace{\bar{V}^k_{f,1}(x_1,a_1^k)-\bar{V}^{\pi_l^k,\pi_{f}^k}_{f,1}(x_1,a_1^k)}_{\cT_4}\nonumber
\end{align}

Similar to the way we proved the regret bound for the leader, we are going to prove the regret bound for the follower.  There are some subtle differences as we need to consider the fact the follower takes action after observing the leader's action.
In order to bound $\cT_3$ and $\cT_4$, we state and prove some base results.

First, similar to Lemma~\ref{lem:leader_q_diff}, we state and prove Lemma~\ref{lem:follower_q_diff} which characterizes the difference between $\langle w_{f,h}^k,\phi(x,a,b)\rangle$ and $Q_{f,h}^{\pi_l,\pi_f}$. In particular, we show that the gap between $\langle \phi(x,a,b), w_{l,h}^k\rangle$ and $\langle \phi(x,a,b), w_{l,h}^{\pi_l,\pi_f}(x,a,b)\rangle$ is upper bounded by expected gap between $V_{f,h+1}^k$ and $V_{f,h+1}^{\pi_l,\pi_f}$ plus the bonus term (Lemma~\ref{lem:follower_q_diff}). In order to prove Lemma~\ref{lem:follower_q_diff}, we state and prove Lemma~\ref{lem:phi_f} which lies in the heart of the proof of regret.  Further, we show that if one use soft-max, the sub-optimality gap can be bounded (Lemma~\ref{lem:close_optimal_f}). Now, recall that we add the bonus term to $\langle \phi(x,a,b), w_{f,h}^k\rangle$ to obtain $Q_{f,h}^k$. Thus, combining all of the above we bound $\cT_3$ using backward induction since the follower also computes the leader's policy. Using Lemma~\ref{lem:follower_q_diff}, we show that the difference between $\bar{V}_{f,h}^k-\bar{V}_{f,h}^{\pi_l,\pi_f}$ can be upper bounded by sum of Martingale differences and sum of the bonus terms (Lemma~\ref{lm:recursion}). Thus, $\cT_4$ can be bounded using Azuma-Hoeffding inequality and elliptical potential lemma.

\subsection{Fomral Proof}
Similar to Lemma~\ref{lem:phi}, we can show the following
\begin{lem}\label{lem:phi_f}
There exists a constant $C_2$ such that for any fixed $p\in (0,1)$, if we let $\mathcal{E}$ be the event that
\begin{align}
\norm{\sum_{\tau=1}^{k-1}\phi_{h}^{\tau}[V_{f,h+1}^{k}(x_{h+1}^{\tau})-\bP_hV_{f,h+1}^{k}(x_h^{\tau},a^{\tau}_h,b_h^{\tau})}_{(\Lambda_h^k)^{-1}}\leq C_2dH\sqrt{\chi}
\end{align}
 $\chi=\log[4(C_1+1)(\log(|\cA||\cB|)+2\log(|\cA|)\log(|\cB|)) dT/p]$, for some constant $C_2$, then $\Pr(\mathcal{E})=1-p/2$. 
\end{lem}

Similar to Lemma~\ref{lem:leader_q_diff}, we can also show the following for the follower as well.

\begin{lem}\label{lem:follower_q_diff}
There exists an absolute constant $\beta=C_1dH\sqrt{\iota}$, $\iota=\log((\log(|\cA||\cB|)+2\log(|\cA|)\log(|\cB|))4dT/p)$, and for any fixed policy $\pi_f$, on the event $\mathcal{E}$ defined in Lemma~\ref{lem:phi_f}, we have 
\begin{align}
\langle \phi(x,a,b),w_{l,h}^k\rangle-Q_{f,h}^{\pi_l,\pi_f}(x,a,b)=\bP_h(V_{f,h+1}^k-V^{\pi_l,\pi_f}_{f,h+1})(x,a,b)+\Delta_h^k(x,a,b)
\end{align}
for some $\Delta_h^k(x,a)$ that satisfies $|\Delta_h^k(x,a,b)|\leq \beta\sqrt{\phi(x,a,b)^T(\Lambda_h^k)^{-1}\phi(x,a,b)}$.
\end{lem}

Similar to Lemma~\ref{lem:close_optimal}, we also obtain the following
\begin{lem}\label{lem:close_optimal_f}
Then, $\tilde{V}_{f,h}^k(x,a)-\bar{V}_{f,h}^{k}(x,a)\leq \dfrac{\log|\cB|}{\alpha}$
\end{lem}
where 
\begin{definition}\label{defn:barvhk_f}
 $\tilde{V}_{f,h}^k(\cdot,a)=\sum_{b}\pi_{f,h}^k(b|\cdot,a)Q_{f,h}^k(\cdot,a,b)]$.
\end{definition}
$\tilde{V}_{f,h}^k(\cdot)$ is the value function corresponds to the greedy-policy with respect to the  $Q$ function given the leader's action. There is a slight difference with Lemma~\ref{lem:close_optimal}. Here, we are bounding the gap for a given leader's action.  



\begin{proof}
Note that
\begin{align}
\bar{V}_{f,h}^{k}(x,a)=\sum_{b}\pi_{f,h}^k(b|x,a)Q_{f,h}^k(x,a,b)\nonumber
\end{align}
where
\begin{align}\label{eq:boltz1}
    \pi_{f,h}^k(b|x,a)=\dfrac{\exp(\alpha_f[Q_{f,h}^k(x,a,b)])}{\sum_{b}\exp(\alpha_f[Q_{f,h}^k(x,a,b)])}
\end{align}
Denote $b_x=\arg\max_{b}Q_{f,h}^k(x,a,b)$

Now, recall from Definition~\ref{defn:barvhk_f} that $\tilde{V}_{f,h}^k(x,a)=Q_{f,h}^k(x,a,b_x)$. Then,
\begin{align}\label{eq:uppb}
     \tilde{V}_{f,h}^k(x,a)-\bar{V}_{f,h}^{k}(x,a)& =Q_{f,h}^k(x,a,b_x) - \sum_{b}\pi_{f,h}^k(b|x,a)Q_{f,h}^k(x,a,b)\nonumber\\
 & =\left(\dfrac{\log(\sum_{b}\exp(\alpha_f(Q_{f,h}^k(x,a,b))))}{\alpha_f}\right)-\sum_{b}\pi_{f,h}^k(b|x,a)Q_{f,h}^k(x,a,b)\nonumber\\
  & \leq \dfrac{\log(|\cB|)}{\alpha_f}
\end{align}
where the last inequality follows from Proposition 1 in \cite{pan2019reinforcement}.
\end{proof}

We are now ready to prove the first part of Lemma~\ref{lem:follower}, i.e., we obtain the bound for $\cT_3$.
\begin{proof}
We prove the lemma by Induction.


First, we prove for the step $H$. 

 Note that $Q_{f,H+1}^k=0=Q_{f,H+1}^{\pi}$.

Under the event in $\mathcal{E}$ as described in Lemma~\ref{lem:phi_f} and \ref{lem:follower_q_diff}, we have 
\begin{align}
& |\langle\phi(x,a,b),w_{f,H}^k\rangle-Q_{f,H}^{\pi}(x,a,b)| \leq  \beta\sqrt{\phi(x,a,b)^T(\Lambda_H^k)^{-1}\phi(x,a,b)}\nonumber
\end{align}
Hence, for any $(x,a,b)$,
\begin{align}
 Q_{f,H}^{\pi}(x,a,b)& \leq \min\{\langle\phi(x,a,b),w_{f,H}^k\rangle+\beta\sqrt{\phi(x,a,b)^T(\Lambda_H^k)^{-1}\phi(x,a,b)},H\}\nonumber\\& 
= Q_{f,H}^k(x,a,b)\nonumber
\end{align}

Hence, from the definition of $\tilde{V}_h^k$, for any $(x,a)$
\begin{align}
\tilde{V}_{H}^k(x,a)& =\max_{b}Q_{f,H}^k(x,a,b)\geq \sum_{b}\pi_{f,h}^k(b|x,a)Q_{f,h}^{\pi}(x,a,b) \nonumber\\
& =\bar{V}_{f,H}^{\pi}(x,a)\nonumber
\end{align}
for any policy $\pi$.  Thus, it also holds for $(\pi_l,\pi^{k,*}_{f})$, the optimal policy for the follower for a given leader's policy $\pi_l$. Hence, from Lemma~\ref{lem:close_optimal_f}, we have for any $a$
\begin{align}
    \bar{V}_{f,H}^{\pi_l,\pi_f^*}(x,a)-\bar{V}_{f,H}^{k}(x,a)\leq \dfrac{\log(|\cB|)}{\alpha_f}\nonumber
\end{align}

Now, suppose that it is true till the step $h+1$ and consider the step $h$. Thus,
\begin{align}
\bar{V}_{f,h+1}^{\pi^k_l,\pi_f^{k,*}}(x,a)-\bar{V}_{f,h+1}^k(x,a)\leq \dfrac{(H-h)\log(|\cB|)}{\alpha_f}
\end{align}

Hence, for a given policy $\pi_l$, we have
\begin{align}\label{eq:diffvalue_f}
    \sum_{a}\pi_{l,h+1}(a|x)(V_{f,h+1}^{\pi_l,\pi_f^*}(x,a)-V_{f,h+1}^{k}(x,a))\leq \dfrac{(H-h)\log(|\cB|)}{\alpha_f}
\end{align}
since $\sum_{a}\pi_{l,h+1}(a|x)=1$. However, in (\ref{eq:diffvalue_f}, the left-hand side is exactly equal to $V_{f,h+1}^{\pi_l,\pi_f^*}(x)-V_{f,h+1}^{k}(x)$. Hence, we have
\begin{align}
    V_{f,h+1}^{\pi_l,\pi_f^*}(x)-V_{f,h+1}^{k}(x)\leq \dfrac{(H-h)\log(|\cB|)}{\alpha_f}
\end{align}


From Lemma~\ref{lem:follower_q_diff} and the above result, we have for any $(x,a,b)$ for any policy $\pi=\{\pi_l,\pi_f\}$
\begin{align}
& Q_{f,h}^{\pi}(x,a,b)  \leq  Q_{f,h}^{k}(x,a,b)+\dfrac{(H-h)\log(|\cB|)}{\alpha}
\end{align}

From the definition of $\tilde{V}_{f,h}^k$, we have
\begin{align}
   &  \sum_{b}\pi_f(b|x,a)Q_{f,h}^{\pi}(x,a,b)  \leq  \max_{b}Q_{f,h}^{k}(x,a,b)+\sum_{b}\pi_f(b|x,a)\dfrac{(H-h)\log(|\cB|)}{\alpha}
\end{align}
Since $\sum_b\pi_f(b|x,a)=1$, and from the Definition of $\tilde{V}_{f,h}^k$, we obtain
\begin{align}
    \bar{V}^{\pi}_{f,h}(x,a)\leq \tilde{V}_{f,h}^k(x,a)+\dfrac{(H-h)\log(|\cB|)}{\alpha_f}
\end{align}
Now, from Lemma~\ref{lem:close_optimal_f}, 
 we have $\tilde{V}_{f,h}^{k}(x,a)-\bar{V}_{f,h}^k(x,a)\leq \dfrac{\log(|\cB|)}{\alpha_f}$. Thus, for any $(x,a)$
\begin{align}
    \bar{V}^{\pi}_{f,h}(x,a)-\bar{V}_{f,h}^k(x,a)\leq \dfrac{(H-h+1)\log(|\cB|)}{\alpha_f}\nonumber
\end{align}
Now, since it is true for any policy $\pi$, it will be true for $\{\pi_l,\pi^{*}_f\}$. From the definition of $\bar{V}^{\pi}_{f,h}$, we have
\begin{align}
    \bar{V}^{\pi_l,\pi_f^*}_{f,h}(x,a_h)-\bar{V}_{f,h}^k(x,a_h)\leq \dfrac{(H-h+1)\log(|\cB|)}{\alpha_f}
\end{align}
Hence, the result follows by summing over $K$ and considering $h=1$. 
\end{proof}

In order to prove the second part of Lemma~\ref{lem:follower} (i.e., in order to bound $\cT_4$), we state and prove a result. 
 
 First, we introduce some  notations. 
 Let
\begin{align}\label{eq:d_martingale}
    & D_{f,h,1}^k=\langle(Q_{f,h}^k(x_{h}^k,a_h^k,\cdot)-Q_{f,h}^{\pi^{k}}(x_{h}^k,a_h^k,\cdot)),\pi_{f,h}^k(\cdot|x_h^k,a_h^k)\rangle-(Q_{f,h}^{k}(x_h^k,a_h^k,b_h^k)-Q_{f,h}^{\pi^k}(x_h^k,a_h^k,b_h^k))\nonumber\\
    & D_{f,h,2}^k=\bP_h(V_{f,h+1}^k-V_{f,h+1}^{\pi^k})(x_h^k,a_h^k,b_h^k)-[V_{f,h+1}^k-V_{f,h+1}^{\pi^k}](x_{h+1}^k)
\end{align}

\begin{lem}\label{lm:recursion}
 On the event defined in $\mathcal{E}$ in Lemma~\ref{lem:phi_f}, we have
 \begin{align}
    \bar{V}_{f,1}^{k}(x_1,a_1^k)-\bar{V}_{f,1}^{\pi^k}(x_1,a_1^k)\leq\sum_{h=1}^{H}(D_{f,h,1}^k+D_{f,h,2}^k)+\sum_{h=1}^{H}2\beta\sqrt{\phi(x_{h}^k,a_{h}^k,b_h^k)^T(\Lambda_h^k)^{-1}\phi(x_{h}^k,a_{h}^k,b_h^k)}\nonumber
\end{align}
\end{lem}
\begin{proof}
By Lemma~\ref{lem:follower_q_diff}, for any $x,h,a,b,k$
\begin{align}
& \langle w_{f,h}^k,\phi(x,a,b)\rangle+\beta\sqrt{\phi(x,a,b)^T(\Lambda_h^k)^{-1}\phi(x,a,b)}-Q_{f,h}^{\pi^k}(x,a,b) \nonumber\\
& \leq \bP_h(V_{f,h+1}^k-V_{f,h+1}^{\pi^k})(x,a,b)+2\beta\sqrt{\phi(x,a,b)^T(\Lambda_h^k)^{-1}\phi(x,a,b)}
\end{align}
Thus, 
\begin{align}\label{eq:q_d}
Q_{f,h}^{k}(x,a,b)-Q_{f,h}^{\pi^k}(x,a,b)\leq \bP_h(V_{f,h+1}^k-V_{f,h+1}^{\pi^k})(x,a,b)+2\beta\sqrt{\phi(x,a,b)^T(\Lambda_h^k)^{-1}\phi(x,a,b)}\nonumber\\
\bP_h(V_{f,h+1}^k-V_{f,h+1}^{\pi^k})(x,a,b)+2\beta\sqrt{\phi(x,a,b)^T(\Lambda_h^k)^{-1}\phi(x,a,b)}-(Q_{f,h}^{k}(x,a,b)-Q_{f,h}^{\pi^k}(x,a,b))\geq 0
\end{align}
Since $\bar{V}_{f,h}^{k}(x,a)=\sum_{b}\pi_{f,h}^k(b|x,a)Q_{f,h}^k(x,a,b)$ and $\bar{V}_{f,h}^{\pi^k}(x,a)=\sum_{b}\pi_{f,h}^k(b|x,a)Q_{f,h}^{\pi^k}(x,a,b)$. 

Thus, from (\ref{eq:q_d}),
\begin{align}\label{eq:recursive}
    & \bar{V}_{f,h}^{k}(x_h^k,a_h^k)-\bar{V}_{f,h}^{\pi^k}(x_h^k,a_h^k)=\sum_{b}\pi_{f,h}^k(b|x_h^k,a_h^k)[Q_{f,h}^{k}(x_{h}^k,a_h^k,b)-Q_{f,h}^{\pi^k}(x_{h}^k,a_h^k,b)]\nonumber\\
    & \leq \sum_{b}\pi_{f,h}^k(b|x_h^k,a_h^k)[Q_{f,h}^{k}(x_{h}^k,a_h^k,b)-Q_{f,h}^{\pi^k}(x_{h}^k,a_h^k,b)]\nonumber\\
    & +2\beta\sqrt{\phi(x_{h}^k,a_{h}^k,b_h^k)^T(\Lambda_h^k)^{-1}\phi(x_{h}^k,a_{h}^k,b_h^k)}+\bP_h(V_{f,h+1}^k-V_{f,h+1}^{\pi^k})(x_{h}^k,a_h^k,b_h^k)-(Q_{f,h}^{k}(x_h^k,a_h^k,b_h^k)-Q_{f,h}^{\pi^k}(x_h^k,a_h^k,b_h^k))
\end{align}

Thus, from (\ref{eq:recursive}), we have
\begin{align}
    \bar{V}_{f,h}^k(x_h^k,a_h^k)-\bar{V}_{f,h}^{\pi^k}(x_h^k,a_h^k)\leq D_{f,h,1}^k+D_{f,h,2}^k+[V_{f,h+1}^k-V_{f,h+1}^{\pi^k}](x_{h+1}^k)+2\beta\sqrt{\phi(x_{h}^k,a_{h}^k,b_h^k)^T(\Lambda_h^k)^{-1}\phi(x_{h}^k,a_{h}^k,b_h^k)}
\end{align}
Hence, by iterating recursively, we have
\begin{align}
    \bar{V}_{f,1}^{k}(x_1,a_1^k)-\bar{V}_{f,1}^{\pi^k}(x_1,a_1^k)\leq\sum_{h=1}^{H}(D_{f,h,1}^k+D_{f,h,2}^k)+\sum_{h=1}^{H}2\beta\sqrt{\phi(x_{h}^k,a_{h}^k,b_h^k)^T(\Lambda_h^k)^{-1}\phi(x_{h}^k,a_{h}^k,b_h^k)}
\end{align}
The result follows.
\end{proof}
Now,, we are ready to obtain the upper bound for $\cT_4$.
\begin{proof}
Note from Lemma~\ref{lm:recursion}, we have
\begin{align}\label{eq:martin_f}
\sum_{k=1}^{K}\bar{V}_{f,1}^{k}(x_1,a_1^k)-\bar{V}_{f,1}^{\pi^k}(x_1,a_1^k)\leq \sum_{k=1}^{K}\sum_{h=1}^{H}(D_{f,h,1}^k+D_{f,h,2}^k)+\sum_{k=1}^{K}\sum_{h=1}^{H}2\beta\sqrt{\phi(x_h^k,a_h^k,b_h^k)^T(\Lambda_h^k)^{-1}\phi(x_h^k,a_h^k,b_h^k)}
\end{align}
We, now, bound the individual terms. First, we show that the first term corresponds to a Martingale difference.

For any $(k,h)\in [K]\times [H]$, we define $\cF_{h,1}^k$ as $\sigma$-algebra generated by the state-action sequences, rewards for leader and follower $\{(x_i^{\tau},a_i^{\tau},b_i^{\tau})\}_{(\tau,i)\in [k-1]\times [H]}\cup \{(x^k_{i},a_i^k,b_i^k)\}_{i\in [h]}$. 

Similarly, we define the $\cF_{h,2}^k$ as the $\sigma$-algebra generated by $\{(x_i^{\tau},a_i^{\tau},b_i^{\tau})\}_{(\tau,i)\in [k-1]\times [H]}\cup \{(x^k_{i},a_i^k,b_i^k)\}_{i\in [h]}\cup\{x_{h+1}^k,a_{h+1}^k\}$. $x_{H+1}^k$ is a null state for any $k\in [K]$. 

A filtration is a sequence of $\sigma$-algebras $\{\cF_{h,m}^k\}_{(k,h,m)\in [K]\times[H]\times[2]}$ in terms of time index
\begin{align}
    t(k,h,m)=2(k-1)H+2(h-1)+m\nonumber
\end{align}
which holds that $\cF_{h,m}^k\subset \cF_{h^{\prime},m^{\prime}}^{k^{\prime}}$ for any $t\leq t^{\prime}$.  Also,
 $\cF_{h,m}^k\subset \cF_{h^{\prime},m^{\prime}}^{k^{\prime}}$ for any $(h,m,k)\leq (h^{\prime},m^{\prime},k^{\prime})$. 

Note from the definitions in (\ref{eq:d_martingale}) that $D_{f,h,1}^k\in \mathcal{F}_{h,1}^k$ and $D_{f,h,2}^k\in \mathcal{F}_{h,2}^k$. Thus, for any $(k,h)\in [K]\times [H]$, 
\begin{align}
    \mathbbm{E}[D_{f,h,1}^k|\cF_{h-1,2}^k]=0, \quad \mathbbm{E}[D_{f,h,2}^k|\cF_{h,1}^k]=0\nonumber
\end{align}
Notice that $t(k,0,2)=t(k-1,H,2)=2(H-1)k$. Clearly, $\cF_{0,2}^k=\cF_{H,2}^{k-1}$ for any $k\geq 2$. Let $\cF_{0,2}^1$ be empty. We define a Martingale sequence
\begin{align}
    M_{f,h,m}^k& = \sum_{\tau=1}^{k-1}\sum_{i=1}^{H}(D_{f,i,1}^{\tau}+D_{f,i,2}^{\tau})+\sum_{i=1}^{h-1}(D_{f,i,1}^{k}+D_{f,i,2}^k)+\sum_{l^{\prime}=1}^{m}D_{f,h,l^{\prime}}^k\nonumber\\
    & =\sum_{(\tau,i,l^{\prime})\in [K]\times[H]\times[2]}D^{\tau}_{f,i,l^{\prime}}
\end{align}
 Clearly, this martingale is adopted to the filtration $\{\cF_{h,m}^k\}_{(k,h,m)\in [K]\times [H]\times [2]}$, and particularly
\begin{align}
    \sum_{k=1}^K\sum_{h=1}^H(D_{f,h,1}^k+D_{f,h,2}^k)=M_{f,H,2}^K
\end{align}

Thus, $M_{f,H,2}^K$ is a Martingale difference satisfying $|M_{f,H,2}^K|\leq 4H$ since $|D_{f,h,1}^k|,|D_{f,h,2}^k|\leq 2H$
From the Azuma-Hoeffding inequality, we have
\begin{align}
\Pr(M_{f,H,2}^K> s)\leq 2\exp(-\dfrac{s^2}{16TH^2})
\end{align}
With probability $1-p/4$ at least,
\begin{align}
\sum_{k}\sum_{h}M_{f,H,2}^K\leq \sqrt{16TH^2\log(4/p)}
\end{align}

Now, we bound the second term. Note that the minimum eigen value of $\Lambda_h^k$ is at least $\lambda=1$ for all $(k,h)\in [K]\times [H]$. By Lemma~\ref{lem:le1}, 
\begin{align}
\sum_{k=1}^{K}(\phi_h^k)^T(\Lambda_h^k)^{-1}\phi_h^k\leq 2\log\left[\dfrac{\det(\Lambda_h^{k+1})}{\det(\Lambda_h^1)}\right]
\end{align}
Moreover, note that $||\Lambda_h^{k+1}||=||\sum_{\tau=1}^{k}\phi_h^k(\phi_h^k)^T+\lambda\bI||\leq \lambda+k$, hence,
\begin{align}
   \sum_{k=1}^{K} (\phi_h^k)^T(\Lambda_h^k)^{-1}\phi_h^k\leq 2d\log\left[\dfrac{\lambda+k}{\lambda}\right]\leq 2d\iota
\end{align}
Now, by Cauchy-Schwartz inequality, we have
\begin{align}
    \sum_{k=1}^{K}\sum_{h=1}^H \sqrt{(\phi_h^k)^T(\Lambda_h^k)^{-1}\phi_h^k}& \leq \sum_{h=1}^{H}\sqrt{K}[\sum_{k=1}^{K}(\phi_h^k)^T(\Lambda_h^k)^{-1}\phi_h^k]^{1/2}\nonumber\\
    & \leq H\sqrt{2dK\iota}
\end{align}
Note that $\beta=C_1dH\sqrt{\iota}$. 

Thus, we have with probability $1-p/2$,
\begin{align}
\sum_{k=1}^{K}V_{f,1}^{k}(x_1^{k},a_1^{k})-V_{f,1}^{\pi_k}(x_1^k,a_1^{k})
\leq  [\sqrt{16TH^2\log(4/p)}+C_4\sqrt{d^3H^3T\iota^2}]
\end{align}
Hence, the result follows. 
\end{proof}

\section{Proof of Lemma~\ref{lem:phi}}
In this section, we prove Lemma~\ref{lem:phi}. First, we provide some supporting result.

\subsection{Supporting Result}
In order to prove Lemma~\ref{lem:phi}, we leverage on the following result which has been proved in \cite{abbasi2011improved}.

\begin{theorem}\label{thm:self_norm}[Concentration of Self-Normalized Process \cite{abbasi2011improved}]
Let $\{\epsilon_t\}_{t=1}^{\infty}$ be a real-valued stochastic process with corresponding filtration $\{\cF_t\}_{t=0}^{\infty}$. Let $\epsilon_t|\cF_{t-1}$ be a zero mean and $\sigma$ sub-Gaussian, i.e., $\bE[\epsilon_t|\cF_{t-1}]=0$, and
\begin{align}
    \forall \zeta\in \Re, \quad \bE[e^{\zeta\epsilon_t}|\cF_{t-1}]\leq e^{\zeta^2\sigma^2/2}.
\end{align}
Let $\{\phi_t\}_{t=1}^{\infty}$ be a $\Re^d$-valued Stochastic process where $\phi_t\in \cF_{t-1} $. Assume $\Lambda_0\in \Re^{d\times d}$ is a positive-define matrix,  let,  $\Lambda_t=\Lambda_0+\sum_{j=0}^{t}\phi_j\phi_j^T\phi_j$. Then for any $\delta>0$ with probability at least $1-\delta$, we have 
\begin{align}
    ||\sum_{s=1}^{t}\phi_s\epsilon_s||_{\Lambda_t^{-1}}^2\leq 2\sigma^2\log\left[\dfrac{\det(\Lambda_t)^{1/2}\det(\Lambda_0)^{-1/2}}{\delta} \right]
\end{align}
\end{theorem}

Further, we also use the following result from \cite{jin2020provably} (Lemma D.4)
\begin{proposition}\label{prop:uniconc}
Let $\{x_{\tau}\}_{\tau=1}^{\infty}$ be a stochastic process on state space $\cS$ with corresponding Filtration $\{\cF_{\tau}\}_{\tau=0}^{\infty}$. Let $\{\phi_{\tau}\}_{\tau=1}^{\infty}$ be a $\mathbbm{R}^d$-valued stochastic process where $\phi_{\tau}\in \cF_{\tau-1}$, and $||\phi_{\tau}||\leq 1$. Let $\Lambda_k=\lambda \bI+\sum_{\tau=1}^{k-1}\phi_{\tau}\phi_{\tau}^T$. Then for any $\delta>0$ with probability $1-\delta$, for all $k\geq 0$, and any $V\in \cV$ so that $\sup_{x}|V(x)|\leq H$, we have
\begin{align}\label{eq:uniconcentration}
    \norm{\sum_{\tau=1}^{k-1}\phi_{\tau}[V(x_{\tau})-\bE[V(x_{\tau}|\cF_{\tau-1})]]}_{(\Lambda_h^k)^{-1}}\leq 4H^2\left[\dfrac{d}{2}\log(\dfrac{k+\lambda}{\lambda})+\log\dfrac{N_{\epsilon}^V}{\delta}\right]+8k^2\epsilon^2/\lambda^2
\end{align}
where $N_{\epsilon}^V$ is the $\epsilon$-covering number for the value function class $\cV$.
\end{proposition}

\subsection{Proof Outline}
Recognizing that (\ref{eq:uniconcentration}) is the same as in the statement of Lemma~\ref{lem:phi}, we only need to compute the $\epsilon$-covering number $N_{\epsilon}^V$ for the value function class of leader. In the following, we compute the $\epsilon$-covering number for value function class (Lemma~\ref{cor:vcovering}).

In order to compute that  we first compute the $\epsilon$-covering number of the individual joint state-action pair value function ($Q$-functions) (Lemma~\ref{lm:qcovering}). Subsequently, we show that if the two $Q$-functions are close, the policies are also close (Lemma~\ref{lem:pi}). Using the above, we compute the $\epsilon$-covering number for the class of marginal $Q$-function ($q_l$) (Lemma~\ref{lem:marginal_covering}).  Finally, again applying the fact that the leader's policies are close as the leader is employing soft-max policy, and $\epsilon$-covering number for the marginal $Q$ function we compute the $\epsilon$-covering number for value function class. Finally, combining Proposition~\ref{prop:uniconc} and Lemma~\ref{cor:vcovering}, we prove Lemma \ref{lem:phi}. 

\subsection{Formal Proof}
In the following we drop the subscript $h$  from $Q_{l,h}^k$, $Q_{f,h}^k$, $V_{l,h}^k$, and $q_{l,h}^k$ for notational simplicity.

We first define the value function. In order to do that, we need to define the $Q$ function class for both the leader and the follower
\begin{definition}
Let  $\mathcal{Q}_m=\{Q|Q(\cdot,\cdot,\cdot)=\min\{w_m^T\phi(\cdot,\cdot,\cdot)+\beta\sqrt{\phi^T(\cdot,\cdot,\cdot)^T\Lambda^{-1}\phi(\cdot,\cdot,\cdot)},H\}\}$ for $m=l,f$, where $\{w\in \R^d|||w||\leq 2H\sqrt{dk/\lambda}\}$,
\end{definition}
Note from Lemma~\ref{lem:w} such class of $Q$ function indeed covers the estimated $Q$-functions we obtain in our algorithm.

We now define the marginal (or,induced) $Q$-function class for the leader
\begin{definition}\label{defn:class_marginal}
Consider the class of function $\hat{\cQ}_l=\{q_l|q_l(\cdot,a)=\langle \pi_f(\cdot|\cdot,a)Q_{l}(\cdot,a,\cdot), \pi_f\in \Pi_{\cB}\rangle\}$
where 
\begin{align*}
   \Pi_{\cB} =\{\pi| \forall b\in \cB, \pi(b|\cdot,\cdot) &= \textsc{Soft-Max}^b_{\alpha_f}((Q_{f}(\cdot,\cdot,\cdot)),
 Q_f \in \cQ_f\},
 \end{align*}
\end{definition}
We can now introduce the class of value function for $V_{l}$.
\begin{definition}\label{defn:classv}
Let $\cV_l=\{V_l|V_l(\cdot)=\bD^{\pi_l,\pi_f}Q_m(\cdot,\cdot,\cdot), \pi_l\in \Pi_{\cA},\pi_f\in\Pi_{\cB}\}$  where 

 \begin{align*}
     \Pi_{\cA} =\{\pi| \forall a\in \cA, \pi(a|\cdot) &= \textsc{Soft-Max}^a_{\alpha_l}((q_{l}(\cdot,\cdot)),
 q_l\in \cQ_l\}
 \end{align*}
\end{definition}
The class of value function $\cV_l$ is parameterized by $w_l$, $w_f$,  and $\Lambda$. It is needless to say we can define the value function for the follower in the similar manner.


\begin{lem}\label{cor:vcovering}
     There exists a $\tilde{V}_l\in \mathcal{V}_l$ parameterized by $(\tilde{w}_l,\tilde{w}_f,\tilde{\beta},\Lambda)$ such that DIST $(V_l,\tilde{V}_l)\leq \epsilon$ where
    \begin{align}
        \rD\rI\rS\rT(V_l,\tilde{V}_l)=\sup_{x}|V_l(x)-\tilde{V}_l(x)|.
    \end{align}
    Let $N_{\epsilon}^{V_l}$ be the $\epsilon$-covering number for the set $\cV_l$, then, 
    \begin{align}\label{eq:vcovering}
    \log N_{\epsilon}^{V_l}\leq d\log\left(1+8H\dfrac{\sqrt{dk}}{\sqrt{\lambda}\epsilon^{\prime}}\right)+d^2\log\left[1+8d^{1/2}\beta^2/(\lambda(\epsilon^{\prime})^2)\right]
    \end{align}
    where $\epsilon^{\prime}=\dfrac{\epsilon}{1+2(\alpha_l+\alpha_f)H+4\alpha_l\alpha_fH^2}$.
\end{lem}

In order to prove the above result, we need to state and prove few results first. 

We, first, obtain the $N^{Q}_{\epsilon}$ covering number for the joint state-action value action $\cQ_j$. Towards this end, we first, introduce some notations.

\begin{definition}\label{defn:epsilon_cover}
Let $\cC_w^{\epsilon}$ be an $\epsilon/2$- cover of the set $\{w\in \R^d|||w||\leq 2H\sqrt{dk/\lambda}\}$ with respect to the 2-norm. Let $\cC^{\epsilon}_{\vA}$ be an $\epsilon^2/4$-cover of the set $\{\vA\in \R^{d\times d}|||\vA||_F\leq d^{1/2}\beta^2\lambda^{-1}\}$ with respect to the Frobenius norm. 
\end{definition}

\begin{lem}\label{lm:qcovering}
\begin{align}
    |\mathcal{C}_{w}^{\epsilon}|\leq (1+8H\sqrt{dk/\lambda}/\epsilon)^d, \quad |\mathcal{C}_{\vA}^{\epsilon}|\leq [1+8d^{1/2}\beta^2/(\lambda\epsilon^2)]^{d^2}
\end{align}
The $\epsilon$-covering number for the set $\cQ_m$, for $m=l,f$, $N^{Q_m}_{\epsilon}$ of the set  $\mathcal{Q}_m$ for $m=l,f$ satisfies the following
\begin{align}
\log N^{Q_m}_{\epsilon}\leq d\log\left(1+\dfrac{8H\sqrt{dk}}{\sqrt{\lambda}\epsilon}\right)+d^2\log[1+8d^{1/2}\beta^2/(\lambda\epsilon)^2]
\end{align}
 The distance metric is the $\infty$-norm, i.e., $\mathrm{dist}(Q_1,Q_2)=\sup_{x,a}|Q_1(x,a)-Q_2(x,a)|$.
\end{lem}
\begin{proof}
For notational simplicity, we represent $\vA=\beta^2\Lambda^{-1}$, and reparamterized the class $\cQ_m$ by $(w_m,\vA)$.  Now,
\begin{align}
dist(Q_1,Q_2)& = \sup_{x,a}|[w_1^T\phi(x,a,b)+\sqrt{\phi^T(x,a,b)\vA_1\phi(x,a,b)}]-
 [w_2^T\phi(x,a,b)+\sqrt{\phi^T(x,a)\vA_2\phi(x,a,b)}]|\nonumber\\
& \leq \sup_{\phi:||\phi||\leq 1}|[w_1^T\phi+\sqrt{\phi^T\vA_1\phi}]-
 [w_2^T\phi+\sqrt{\phi^T\vA_2\phi}]|\nonumber\\
& \leq \sup_{\phi:||\phi||\leq 1}|(w_1-w_2)^T\phi|+\sup_{\phi:||\phi||\leq 1}\sqrt{|\phi^T(\vA_1-\vA_2)\phi|}\nonumber\\
& =||w_1-w_2||+\sqrt{||\vA_1-\vA_2||}\leq ||w_1-w_2||+\sqrt{||\vA_1-\vA_2||_F}
\end{align}
where the second-last inequality follows from the fact that $|\sqrt{x}-\sqrt{y}|\leq \sqrt|x-y|$. For matrices $||\cdot||$, and $||\cdot||_F$ denote matrix operator norm and the Frobenius norm respectively. 

Recall that $\cC_w$ is an $\epsilon/2$- cover of the set $\{w\in \R^d|||w||\leq 2H\sqrt{dk/\lambda}\}$ with respect to the 2-norm. Also recall that $\cC_{\vA}$ be an $\epsilon^2/4$-cover of the set $\{\vA\in \R^{d\times d}|||\vA||_F\leq d^{1/2}\beta^2\lambda^{-1}\}$. Thus, from Lemma~\ref{lem:cover_number},
\begin{align}
|\mathcal{C}_{w}^{\epsilon}|\leq (1+8H\sqrt{dk/\lambda}/\epsilon)^d, \quad |\mathcal{C}_{\vA}^{\epsilon}|\leq [1+8d^{1/2}\beta^2/(\lambda\epsilon^2)]^{d^2}\nonumber
\end{align}
For any $Q_m\in \mathcal{Q}_m$, there exists a $\tilde{Q}_m$ parameterized by $(w_{2},\vA_2)$ where  $w_2\in \cC_w^{\epsilon}$ and $\vA_2\in \cC_{\vA}^{\epsilon}$ such that $\mathrm{dist}(Q_m,\tilde{Q}_m)\leq \epsilon$. Hence, $N^{Q_m}_{\epsilon}\leq |\mathcal{C}^{\epsilon}_{w}||\mathcal{C}^{\epsilon}_{\vA}|$, which gives the result since $\log(\cdot)$ is an increasing function.
\end{proof}
We now show that when $Q$-functions are close soft-max policy based on the $Q$-functions are close. In fact the soft-max policy is at most $2\alpha$- Lipschitz.

\begin{lem}\label{lem:pi}
Suppose that $\pi$ is the soft-max policy (temp. coefficient $1/\alpha$) corresponding to the joint state-action value function $Q$-  i.e., $\forall b\in \cB$
\begin{align}
    \pi(b|\cdot,a)=\textsc{Soft-Max}^b_{\alpha_f}(Q^k_{f}(\cdot,a,\cdot)).\nonumber
\end{align}
 $\tilde{\pi}$ is the soft-max policy vector with the same temp. coefficient $1/\alpha$ corresponding to the composite $Q$-function $\tilde{Q}_m$, i.e, $\forall a\in \cA$,
 \begin{align}
     \tilde{\pi}(b|\cdot,a)=\textsc{Soft-Max}^b_{\alpha_f}(\tilde{Q}_{f}(\cdot,a,\cdot)).\nonumber
 \end{align}then, for any state $x$,
\begin{align}
||\pi(\cdot|x,a)-\tilde{\pi}(\cdot|x,a)||_1\leq 2\alpha_f\epsilon^{\prime}
\end{align}
where $\pi(\cdot|x,a)=\{\pi(b|x,a)\}_{b\in \cB}$ and $\tilde{\pi}(\cdot|x,a)=\{\tilde{\pi}(b|x,a)\}_{b\in \cB}$ 
when $\mathrm{dist}(Q_{f}^k,\tilde{Q}_{f})\leq \epsilon^{\prime}$. 
\end{lem}

\begin{proof}
Let $\mathrm{Exp}^{\alpha}(P)$ be a soft-max  corresponding to the vector $P$, i.e., the $i$-th component of $\mathrm{Exp}^{\alpha}(P)$ is
\begin{align}
    \dfrac{\exp(\alpha P_i)}{\sum_{i}\exp(\alpha P_i)}.\nonumber
\end{align}
Note from Theorem 4.4 in \cite{epasto2020optimal} then, we have
\begin{align}\label{eq:use_result}
    ||\mathrm{Exp}^{\alpha}(P_1)-\mathrm{Exp}^{\alpha}(P_2)||_1\leq 2\alpha||P_1-P_2||_{\infty}
\end{align}
for two vectors $P_1$ and $P_2$. 

Now note that in our case for a given state $x$, $\pi$ is equivalent to $\mathrm{Exp}^{\alpha_f}(Q_{f}^k(x,a,\cdot))$, and $\tilde{\pi}$ is equivalent to $\mathrm{Exp}^{\alpha_f}(\tilde{Q}_f(x,a,\cdot))$. Then from (\ref{eq:use_result}) and the fact that $\mathrm{dist}(Q_{f}^k,\tilde{Q}_f)\leq \epsilon^{\prime}$  we have 
\begin{align}
    ||\pi(\cdot|x,a)-\tilde{\pi}(\cdot|x,a)||_1\leq 2\alpha_f\epsilon^{\prime}
\end{align}
Hence, the result follows.
\end{proof}
Now, using the above result, 
we now compute the $\epsilon$-covering number for the marginal $q$-function (Definition~\ref{defn:class_marginal}). 

\begin{lem}\label{lem:marginal_covering}
If $||Q_l-\tilde{Q}_l||\leq \epsilon^{\prime}$, and $||Q_f-\tilde{Q}_f||\leq \epsilon^{\prime}$, then $||q_l-\tilde{q}_l||\leq \epsilon^{\prime}+2\alpha_f\epsilon^{\prime}H$ where $q_l,\tilde{q}_l\in \hat{\cQ}_l$, $Q_m,\tilde{Q}_m\in \cQ_m$, for $m=l,f$.
\end{lem}
\begin{proof}
Consider $Q_f$ and $\tilde{Q}_f$ such that $||Q_f-\tilde{Q}_f||_{\infty}\leq \epsilon^{\prime}$. Hence, from Lemma~\ref{lem:pi}, $||\pi-\tilde{\pi}||_1\leq 2\alpha_f\epsilon^{\epsilon}$.

Consider $Q_l$ and $\tilde{Q}_l$ such that $||Q_l-\tilde{Q}_l||_{\infty}\leq \epsilon^{\prime}$.

Now, 
\begin{align}
    & |q_l(x,a)-\tilde{q}_l(x,a)|=|\sum_{b}\pi(b|x,a)Q_{l}(x,a,b)-\sum_{b}\tilde{\pi}(b|x,a)\tilde{Q}_l(x,a,b)|\nonumber\\
    & \leq |\sum_{b}\pi(b|x,a)(Q_l(x,a,b)-\tilde{Q}_l(x,a,b))|+|\sum_b\pi(b|x,a)\tilde{Q}_l(x,a,b)-\sum_{b}\tilde{\pi}(b|x,a)\tilde{Q}_l(x,a,b)|\nonumber\\
    & \leq \epsilon^{\prime}+||\pi(\cdot|x,a)-\tilde{\pi}(\cdot|x,a)||_1||\tilde{Q}_l||_{\infty}\nonumber\\
    & \leq \epsilon^{\prime}+2\alpha_f\epsilon^{\prime}H
\end{align}
Hence, the result follows.
\end{proof}

Now, we show that if $Q$-functions are close, the leaders' policies are also close based on those $Q$-functions. 

\begin{lem}\label{lem:pi_q_marginal}
Suppose that $\pi_l$ is the soft-max policy (temp. coefficient $1/\alpha_l$) corresponding to the marginal state-action value function $q$-  i.e., $\forall a\in \cA$
\begin{align}
    \pi_l(a|\cdot)=\textsc{Soft-Max}^a_{\alpha_l}(q_{l}(\cdot,\cdot)).\nonumber
\end{align}
 $\tilde{\pi}_l$ is the soft-max policy vector with the same temp. coefficient $1/\alpha_l$ corresponding to the marginal $q$-function $\tilde{q}_l$, i.e, $\forall a\in \cA$,
 \begin{align}
     \tilde{\pi}_l(a|\cdot)=\textsc{Soft-Max}^a_{\alpha_l}(\tilde{q}_{l}(\cdot,\cdot)).\nonumber
 \end{align}then, for any state $x$,
\begin{align}
||\pi_l(\cdot|x)-\tilde{\pi}_l(\cdot|x)||_1\leq 2\alpha_l\epsilon^{\prime}(\epsilon^{\prime}+2\alpha_f\epsilon^{\prime}H)
\end{align}
when $\mathrm{dist}(Q_{f}^k,\tilde{Q}_{f})\leq \epsilon^{\prime}$, $\mathrm{dist}(Q_{l}^k,\tilde{Q}_{l}^k)\leq \epsilon^{\prime}$. 
\end{lem}
\begin{proof}
The proof follows the same steps as in Lemma~\ref{lem:pi}. Now, $||q_l-\tilde{q}_l||\leq \epsilon^{\prime}+2\alpha_f\epsilon^{\prime}H$ (from Lemma~\ref{lem:marginal_covering}). Hence, the result follows.
\end{proof}

\begin{lem}\label{lem:vclose}
There exists $\tilde{V}_m\in \cV_m$ such that
\begin{align}\label{eq:vclose}
\mathrm{DIST}(V_{m}^k,\widetilde{V}_{m})\leq \epsilon^{\prime}+2\alpha_f\epsilon^{\prime}H+2\alpha_l\epsilon^{\prime}H(\epsilon^{\prime}+2\alpha_f\epsilon^{\prime}H) 
\end{align} 
where  $\mathrm{dist}(\tilde{Q}_m,Q_m)\leq\epsilon^{\prime}$, $\tilde{Q}_m\in\cQ_m$ for all $j$;
\begin{align}
    \widetilde{V}_m(\cdot)=\sum_{a}[\tilde{\pi}(a|\cdot)\tilde{q}_l(\cdot\cdot)],\nonumber
\end{align}
\begin{align}
    \tilde{\pi}_l(a|\cdot)=\textsc{Soft-Max}^a_{\alpha_l}(\tilde{q}_{l}(\cdot,a)),\quad  \forall a\in \cA\nonumber
\end{align}
.
\end{lem}
\begin{proof}
For any $x$,
\begin{align}
& V_{l}^k(x)-\widetilde{V}_l(x)\nonumber\\
    & = |\sum_{a}\pi(a|x)q_{l}^k(x,a)-\sum_{a}\tilde{\pi}(a|x)\tilde{q}_{l}(x,a)|\nonumber\\
    & =|\sum_{a}\pi(a|x)q_{l}^k(x,a)-\sum_{a}\pi(a|x)\tilde{q}_{l}(x,a)+\sum_{a}\pi(a|x)\tilde{q}_l(x,a)-\sum_{a}\tilde{\pi}(a|x)\tilde{q}_l(x,a)|\nonumber\\
    & \leq |\sum_{a}\pi(a|x)q_{l}^k(x,a)-\sum_{a}\pi(a|x)\tilde{q}_{l}(x,a)|+|\sum_{a}\pi(a|x)\tilde{q}_l(x,a)-\sum_{a}\tilde{\pi}(a|x)\tilde{q}_l(x,a)|\nonumber\\
    & \leq \epsilon^{\prime}+2\alpha_f\epsilon^{\prime}H+||\pi(\cdot|x)-\tilde{\pi}(\cdot|x)||_1||\tilde{q}_l(x)||_{\infty}\nonumber\\
    & \leq \epsilon^{\prime}+2\alpha_f\epsilon^{\prime}H+2\alpha_l\epsilon^{\prime}H(\epsilon^{\prime}+2\alpha_f\epsilon^{\prime}H)
\end{align}
where we use the fact that $\mathrm{dist}(q_{l}^k,\tilde{q}_l)\leq \epsilon^{\prime}+2\alpha_f\epsilon^{\prime}H$, and $\sum_{a}\pi(a|x)= 1$ for the first term and the Holder's inequality in the second  term for the second last inequality. For the last inequality, we use Lemma~\ref{lem:pi}, and the fact that $\tilde{q}_l(x,a)\leq H$ for any $(x,a)$.
 Hence, we have the result.
\end{proof}
 Assuming that $\epsilon^{\prime}\leq 1$, we have 
 $\mathrm{DIST}(V_m^k,\tilde{V}_m)\leq \epsilon^{\prime}+2H(\alpha_f+\alpha_l)\epsilon^{\prime}+4\alpha_l\alpha_fH^2\epsilon^{\prime}$. Now, we are ready to prove Lemma~\ref{cor:vcovering}.
 
\begin{proof}
Fix an $\epsilon$.
Let $\epsilon^{\prime}=\dfrac{\epsilon}{1+2(\alpha_l+\alpha_f)H+4\alpha_l\alpha_fH^2}$, then from Lemma~\ref{lem:vclose},  we have DIST$(V_{l}^k,\widetilde{V}_l)\leq\epsilon$. Thus, we only need to find parameters in the  $\epsilon^{\prime}$-covering of the $Q$-functions as described in Lemma~\ref{lm:qcovering} in order to obtain $\epsilon$-close value function.

Recall the Definition~\ref{defn:epsilon_cover}. Then, there exists $\tilde{w}_l,\tilde{w}_f\in \cC_w^{\epsilon^{\prime}}$ such that
$||\tilde{w}_l-w_l||\leq \dfrac{\epsilon^{\prime}}{2}$, $||\tilde{w}_f-w_f||\leq \dfrac{\epsilon^{\prime}}{2}$. Further, there exists $\vA_2\in \cC_{\vA}^{\epsilon^{\prime}}$   such that $||\vA-\tilde{\vA}||_F\leq \dfrac{\epsilon^{\prime 2}}{4}$, $\vA=\beta^2(\Lambda^k)^{-1}$, $\tilde{\vA}=\beta^2(\tilde{\Lambda})^{-1}$, for some $\tilde{\Lambda}$. Then we obtain $\tilde{Q}_m$ parameterized by $(\tilde{w}_m,\beta,\tilde{\Lambda})$ for $m=l,f$, such that $\mathrm{dist}(Q_m,\tilde{Q}_m)\leq \epsilon^{\prime}$ (by Lemma~\ref{lm:qcovering}).

  Now, from Lemma~\ref{lem:vclose}, we have DIST$(V_{l}^k,\tilde{V}_l)\leq\epsilon$. Hence, there exists $\tilde{V}_l$ parameterized by $\tilde{w}_l,\tilde{w}_f,\tilde{\vA}$, such that Dist($\tilde{V}_l,V_l^k)\leq \epsilon$. Hence, $N_{\epsilon}^{V_l}\leq |\cC_w^{\epsilon^{\prime}}||\cC_{\vA}^{\epsilon^{\prime}}|$. Thus, from Lemma~\ref{lm:qcovering} ,  the $\epsilon$-covering number $N_{\epsilon}^{V_l}$ for the set $\cV_l$ satisfies the following
\begin{align}
    \log N^{V_l}_{\epsilon}\leq d\log\left(1+8H\dfrac{\sqrt{dk}}{\sqrt{\lambda}\epsilon^{\prime}}\right)+d^2\log\left[1+8d^{1/2}\beta^2/(\lambda(\epsilon^{\prime})^2)\right)]. \nonumber
\end{align}

Hence, the result follows.
\end{proof}

From Lemma~\ref{cor:vcovering}, note that we need $\epsilon^{\prime}$ covering for the $Q$-functions where $\epsilon^{\prime}=\dfrac{\epsilon}{1+2(\alpha_l+\alpha_f)H+4\alpha_l\alpha_fH^2}$ if we need to bound DIST $(V_j,\tilde{V}_j)$ by $\epsilon$. 

Now, we are ready to prove Lemma~\ref{lem:phi}. 

\begin{proof}
By Lemma~\ref{cor:vcovering}, we know that there exists $\tilde{V}_j$ in the $\epsilon$-covering for  $\cV_j$ such that for every $x$,
\begin{align}\label{eq:delv}
    V_l(x)=\tilde{V}_l(x)+\Delta V(x)
\end{align}
where $\sup_{x}\Delta V(x)\leq \epsilon$. 

Hence, 
\begin{align}
    \norm{\sum_{\tau=1}^{k}\phi^{\tau}(V_l(x_{\tau})-\mathbbm{E}[V_l(x_{\tau})|\mathcal{F}_{\tau-1}])}^2_{(\Lambda^k)^{-1}}& \leq  
2\norm{\sum_{\tau=1}^{k}\phi^{\tau}(\tilde{V}_l(x_{\tau})-\mathbbm{E}[\tilde{V}_l(x_{\tau})|\mathcal{F}_{\tau-1}])}^2_{(\Lambda^k)^{-1}}\nonumber\\& +
2\norm{\sum_{\tau=1}^{k}\phi^{\tau}(\Delta V(x_{\tau})-\mathbbm{E}[\Delta V(x_{\tau})|\mathcal{F}_{\tau-1}])}^2_{(\Lambda^k)^{-1}}
\end{align}
The last expression is bounded by $\dfrac{8k^2\epsilon^2}{\lambda}$. 

Now, we bound the first term. Note from Lemma~\ref{cor:vcovering} that in order to obtain $\tilde{V}_l$ which satisfies (\ref{eq:delv}), we need to obtain  we need $N^{V_l}_{\epsilon}$ number of elements to obtain such $(\tilde{w}_l,\tilde{w}_f,\beta,\tilde{\Lambda})$.  Such $\tilde{V}_l$ is independent of samples. Hence, we can use the  self-normalization lemma (Theorem~\ref{thm:self_norm}). From Proposition~\ref{prop:uniconc} we obtain
\begin{align}\label{eq:inequal}
    \norm{\sum_{\tau=1}^{k}\phi^{\tau}(\tilde{V}_l(x_{\tau})-\mathbbm{E}[\tilde{V}_l(x_{\tau})|\mathcal{F}_{\tau-1}])}^2_{(\Lambda^k)^{-1}}\leq 2H^2\left[d\log\left(\dfrac{k+\lambda}{\lambda}\right)+\log\left(\dfrac{N_{\epsilon}^{V_l}}{\delta}\right)\right]
\end{align}
where $N^V_{\epsilon_l}$ is upper bounded in (\ref{eq:vcovering}).   $\beta$ is equal to $C_1dH\sqrt{\iota}$ for some constant $C_1$, and $\iota=\log((\log(|\cA||\cB|)+\log(|\cB|)\log(|\cA|)4dT/p)$.   We obtain from (\ref{eq:inequal})
\begin{align}\label{eq:inequal2}
& \norm{\sum_{\tau=1}^{k}\phi^{\tau}(\tilde{V}_l(x_{\tau})-\mathbbm{E}[\tilde{V}_l(x_{\tau})|\mathcal{F}_{\tau-1}])}^2_{(\Lambda^k)^{-1}}\leq\nonumber\\ &
    4H^2\left[\dfrac{d}{2}\log\left(\dfrac{k+\lambda}{\lambda}\right)+d\log\left(1+\dfrac{8H\sqrt{dk}}{\epsilon^{\prime}\sqrt{\lambda}}\right)+d^2\log\left(1+\dfrac{8d^{1/2}\beta^2}{\epsilon^{\prime2}\lambda}\right)+\log\left(\dfrac{4}{p}\right)\right]
\end{align}
where $\epsilon^{\prime}=\dfrac{\epsilon}{1+2(\alpha_l+\alpha_f)H+4\alpha_l\alpha_fH^2}$.
Set $\epsilon=\dfrac{1}{k}$ 
, $\lambda=1$. Thus, $\epsilon^{\prime}=\dfrac{1}{(1+2(\alpha_l+\alpha_f)H+4\alpha_l\alpha_fH^2)k}$. Plugging in the above,  and putting $\alpha_l=\dfrac{\log(|\cA|)K}{2H}$, and $\alpha_f=\dfrac{\log(|\cB|)K}{2H}$,we obtain from (\ref{eq:inequal2})
\begin{align}
& ||\sum_{\tau=1}^{k}\phi^{\tau}(\tilde{V}_l(x_{\tau})-\mathbbm{E}[\tilde{V}_l(x_{\tau})|\mathcal{F}_{\tau-1}])||^2_{\Lambda_k^{-1}}\leq
C_2H^2d^2\log\left(\dfrac{4(C_1+1)(\log(|\cA||\cB|)+2\log(|\cA|)\log(|\cB|))dT}{p}\right)
\end{align}
for some constant $C_2$. Hence, the result follows. 
\end{proof}
\section{Supporting Results}\label{sec:supporting_Results}
The following result is shown in \cite{abbasi2011improved} and in Lemma D.2 in \cite{jin2020provably}.
\begin{lem}\label{lem:le1}
Let $\{\phi_t\}_{t\geq 0}$ be a sequence in $\Re^d$ satisfying $\sup_{t\geq 0}||\phi_t||\leq 1$. For any $t\geq 0$, we define $\Lambda_t=\Lambda_0+\sum_{j=0}^{t}\phi_j\phi_j^T\phi_j$. Then if the smallest eigen value of $\Lambda_0$ be at least $1$, we have
\begin{align}
   \log\left[\dfrac{\det(\Lambda_h^{k+1})}{\det(\Lambda_h^1)}\right]\leq \sum_{k=1}^{K}(\phi_h^k)^T(\Lambda_h^k)^{-1}\phi_h^k\leq 2\log\left[\dfrac{\det(\Lambda_h^{k+1})}{\det(\Lambda_h^1)}\right]
\end{align}
\end{lem}

The next result characterizes the covering number of an Euclidean ball (Lemma 5.2 in \cite{vershynin2010introduction}). 
\begin{lem}\label{lem:cover_number}[Covering Number of Euclidean Ball]
For any $\epsilon>0$, the $\epsilon$-covering number of the Euclidean ball in $\R^d$ with radius $R$ is upper bounded by $(1+2R/\epsilon)^d$.
\end{lem}
\section{Why does the Greedy Policy fail?}
\label{sec:nogreedy}
In this section, using an example we show that the greedy-policy is not Lipschitz. Further, we illustrate that (for leader) we can not use the greedy-policy for the follower based on the joint $Q$-function to obtain the $\epsilon$-close covering for the leader's value function. In the similar realm, we can construct an example where the greedy-policy based on the leader's marginal $q$-function would not provide $\epsilon$-covering number for the follower's value function.  

Consider the following toy-example: 
\begin{example}\label{eg:greedy_fails}
We basically show that we can not obtain $\epsilon$-covering number for marginal $Q$-function ($q_l$) for the leader if the follower's action is based on the greedy policy. Suppose that the cardinality of the action space $|\cB|$, $|\cA|$ are $2$. 
\begin{table*}[!h]
\caption{The $Q$-table for leader and follower, first number are for leader, second numbers are for follower}
\label{table:staalgos}
\vspace{-0.1in}
\begin{center}
\begin{small}
\begin{sc}
\begin{tabular}{ccc}
\toprule
Actions & $b_1$ & $b_2$\\
\midrule
  $a_1$  & $M-\epsilon$,$1-\epsilon/2$ & $0$,$1+\epsilon/2$ \\
  $a_2$ & $0,1$ & $0, 1-\epsilon/2$ \\
\bottomrule
\end{tabular}

\end{sc}
\end{small}
\end{center}
\end{table*}
Further, suppose that $Q$-table is obtained as in Table~\ref{table:staalgos}. Now, consider $\epsilon$-close values  ($\tilde{Q}$) are obtained as in Table~\ref{table:epsilon}.
\begin{table*}[!h]
\caption{The $\epsilon$ close values $\tilde{Q}$-table for leader and follower. The first numbers are for leader, second numbers are for follower}
\label{table:epsilon}
\vspace{-0.1in}
\begin{center}
\begin{small}
\begin{sc}
\begin{tabular}{ccc}
\toprule
Actions & $b_1$ & $b_2$\\
\midrule
  $a_1$  & $M$,$1+\epsilon/2$ & $0$,$1-\epsilon/2$ \\
  $a_2$ & $0,1+\epsilon/2$ & $0, 1-\epsilon/2$ \\
\bottomrule
\end{tabular}

\end{sc}
\end{small}
\end{center}
\end{table*}
Now, we will be able to show how the marginal $q_l$ values change if the policy is greedy. For example for $a_1$, if the follower's policy would have been greedy, $q_l(x,a_1)=M$ for Table~\ref{table:staalgos}, and $q_l(x,a_1)=0$ for values in Table~\ref{table:epsilon}. Thus, slight changes in the values of $Q$ can lead to a drastic changes in the values of the marginal $q$ values. Also note that $M$ is an arbitrary number, thus, such change can be arbitrary. 

If the policy is soft-max, we can easily show that $q_l$ values are also close.
\end{example}

\section{Why does the approach proposed in \cite{zhong2021can} fail?}
 In \cite{zhong2021can}, at every step they find a stackelberg equilibrium using an oracle for an $\epsilon$-approximated $Q$ function for the leader and known rewards of the follower (they consider myopic follower with known rewards). In particular, \cite{zhong2021can} proposed an approach where they truncate $w_{l,h}^k$, $\Lambda_h^k$ to  $\epsilon$-close value of $w_l$, and $\Lambda$ respectively. Subsequently, they obtained equilibrium policies using these $\epsilon$-close values.
 Then the proposed algorithm uses the above equilibrium strategy attained using $\epsilon$-close values as their policies.  Since these $\epsilon$-close values are predetermined using the $\epsilon$-covering set of $w$ and $\Lambda$ (as we have described the $\epsilon$-covering set in Lemma ~\ref{cor:vcovering}), hence, one can apply uniform concentration lemma for the leader \cite{zhong2021can} with error of at most $\epsilon$. 
 
 However, we can not apply the similar trick since here the followers are non-myopic, and the rewards are unknown. In particular, in \cite{zhong2021can}, the followers' $Q$-functions are not required to be estimated as the rewards are known. Thus, the followers' policies are fixed given the action of the leader. However, in our case, as shown in Example~\ref{eg:greedy_fails}, slight change in the $Q$-function can lead to a change in the action for the follower. Thus, an oracle which computes Stackelberg equilibrium, it would also gives a different action which in turn would make the marginal $Q$-function for the leader arbitrary large as shown in Example~\ref{eg:greedy_fails}.  Thus, one can not show that the log $\epsilon$-covering number for the leader's value functions scales with $\log(K)$.
 
 \section{Proof of Corollary~\ref{cor:ccse}}
 First, consider the policy-- $(\tilde{\pi}_l,\tilde{\pi}_f)$ where both the leader and the follower pick any $k\in [K]$ randomly and then follow the policy $\pi_l^k,\pi_f^k$ given by the Algorithms 1 and 2 at episode $k$.
 
 Note from Theorem~\ref{thm:episodic} that 
 \begin{align}
    & \dfrac{1}{K}\left(\sum_{k}\sum_{a_1^k}\pi_l^k(a_1^k|x_1)(\bar{V}^{\pi_l^k,\pi_f^{k,*}}_{f,1}(x_1,a_1^k)-\bar{V}_{f,1}^{\pi_l^k,\pi_f^k}(x_1,a_1^k)\right)\leq \tilde{\cO}(\sqrt{d^3H^4})/\sqrt{K}\nonumber\\
     & V_{f,1}^{\tilde{\pi}_l,\pi_f^*}(x_1)-V_{f,1}^{\tilde{\pi_l},\tilde{\pi_f}}(x_1)\leq \tilde{\cO}(\sqrt{d^3H^4})/\sqrt{K}
 \end{align}
 By selecting $K=\tilde{\cO}(d^3H^4/\epsilon^2)$ we have the result.
 
 Similarly, for the leader we have
 \begin{align}
     & \dfrac{1}{K}\left(\sum_{k}\bar{V}^{\pi_l^{k,*},\pi_f^{k}}_{l,1}(x_1)-\bar{V}_{l,1}^{\pi_l^k,\pi_f^k}(x_1)\right)\leq \tilde{\cO}(\sqrt{d^3H^4})/\sqrt{K}\nonumber\\
     & 
     V_{l,1}^{\pi_l^*,\tilde{\pi}_f}(x_1)-V_{l,1}^{\tilde{\pi_l},\tilde{\pi_f}}(x_1)\leq \tilde{\cO}(\sqrt{d^3H^4})/\sqrt{K}
 \end{align}
 Again by plugging in $K=\tilde{\cO}(d^3H^4/\epsilon^2)$ we have the result.

\end{document}